\documentclass{article}
\usepackage{arxiv}
\usepackage[T1]{fontenc}
\usepackage[utf8]{inputenc}
\usepackage{lmodern,microtype,amsmath,amssymb,amsthm}
\usepackage{graphicx}
\usepackage{booktabs,tabularx,array,multirow,longtable,caption}
\usepackage{xcolor}
\usepackage{enumitem}
\usepackage{tikz,pgfplots}
\usetikzlibrary{arrows.meta,positioning,calc,fit,backgrounds,shapes.geometric}
\usepgfplotslibrary{groupplots}
\pgfplotsset{compat=1.18}
\usepackage{xurl}
\usepackage{placeins}
\usepackage{algorithm}
\usepackage{algpseudocode}
\usepackage[numbers,sort&compress]{natbib}
\usepackage{hyperref}
\usepackage{cleveref}
\hypersetup{colorlinks=true, linkcolor=blue, urlcolor=cyan, citecolor=blue}
\setlist{nosep,leftmargin=1.4em}
\newcommand{\code}[1]{\texttt{\small\detokenize{#1}}}
\newcolumntype{L}[1]{>{\raggedright\arraybackslash}p{#1}}
\newcolumntype{Y}{>{\raggedright\arraybackslash}X}
\definecolor{chartfront}{HTML}{2A78D6}
\definecolor{chartchosen}{HTML}{EB6834}
\definecolor{chartother}{HTML}{C9C8C3}
\definecolor{chartmuted}{HTML}{52514E}
\definecolor{chartgrid}{HTML}{ECEBE8}
\definecolor{boxfill}{HTML}{F3F6F8}
\newcommand{\judgeHR}{91.5}
\newcommand{\judgeHRa}{92.1}
\newcommand{\judgeHRb}{89.0}
\newcommand{\judgeCost}{\$33.97}
\newcommand{\judgeCostNum}{33.97}
\newcommand{\judgeRatio}{5.9}
\newcommand{\judgeOverOff}{8.4}
\newcommand{\judgeOverOn}{15.7}
\newcommand{\judgeOverHind}{3.7}
\newcommand{\judgeWithin}{38}

\newcommand{\judgeFivetran}{92.1}
\newcommand{\judgeFivetranOverOn}{27.4}
\newcommand{\judgeFivetranOverOff}{6.2}

\title{No Fixed Configuration Is Best:\\ Context- and Cost-Aware Tuning of LLM Pipelines,\\ with an Application to\\ Schema Matching and Mapping}

\author{
  Gokhan Kirlik \\ Tessera Labs \And
  Xiuwen Zheng \\ Tessera Labs \And
  Anirudh Sriram \\ Tessera Labs
}

\renewcommand{\shorttitle}{Context- and Cost-Aware Tuning of LLM Pipelines}

\begin{document}
\maketitle

\begin{abstract}
The numeric settings of a multi-stage LLM pipeline (a compound AI system) are
the usual target of tuning. We tuned all of them for an LLM pipeline for schema
matching and mapping, with accuracy and cost as joint objectives, and we show what this buys,
what it cannot buy, and where the remaining accuracy is. First, tuning buys
cost and a little accuracy. On 35 tasks, with whole groups of tasks held out
from tuning, one tuned configuration gains 2.9 points of top-1 accuracy (HR@1) over the
default at 18\% more cost, and another keeps the accuracy of the default at
32\% lower cost. Second, the accuracy depends on a
choice that numeric tuning cannot reach. Restricting the search for each source
column to one assigned target table changes HR@1 by $+30$ points on a MIMIC-III
to OMOP task and by $-52$ points on a task built from database views, more than
numeric tuning gains on nine of ten test tasks. Neither setting is right for
all tasks: on the larger set of 41 tasks used for the remaining results, mean
HR@1 is 75.7\% with the restriction (the default) and 83.1\% without it. Third, choosing the setting per task is not enough
either: even an oracle that picks the better setting for each task reaches
only 87.8\% on the 41 tasks, because the
better setting differs between the columns of one task. The pipeline therefore
runs both settings, and an LLM judge chooses among their candidate answers on the
26\% of columns on which they disagree. This reaches \judgeHR\% HR@1 on the 41
tasks, at \judgeRatio{} times the cost of the default, about one cent per column.
\end{abstract}


\section{Introduction}\label{sec:intro}
Systems built on large language models are rarely a single model call. A
typical system retrieves evidence, narrows a set of candidates, asks a model to
rank or generate, and then checks the answer, and each of these stages exposes
settings: how many candidates to keep, how much metadata to show the model,
whether to make a second call, when to search further. The settings govern
accuracy and cost at the same time. In practice a pipeline is released with one
configuration, and the usual way to improve it is to tune the settings on more
labeled tasks.

We did this for an LLM pipeline for schema matching and mapping, which finds
for each column of a source schema the corresponding column of a target schema
and, where needed, the rule that computes it. We tuned every numeric
setting of every stage, with accuracy and cost as joint objectives; by numeric
settings we mean thresholds, pool sizes and the amount of metadata in a prompt,
and also the switches that turn a single stage on or off, such as a
verification call. Tuning paid off in cost, and only a little in accuracy. This
paper shows what tuning buys, and asks where the rest of the accuracy is. The
answer has three parts, and each part raises the question that the next one
answers.

The first part is that tuning the numeric settings buys cost and a little
accuracy. On 35 tasks, with whole groups of related tasks held out from tuning,
the tuned configurations form a range. At one end, tuning for accuracy alone
raises mean top-1 accuracy (HR@1) from 76.4\% to 79.3\%, for 18\% more cost.
At the other end, the configuration tuned with cost in the objective keeps the
accuracy of the default (76.8\% against 76.4\%) at 32\% lower cost. A third less cost at no
loss of accuracy is a useful result. It also leaves a puzzle. Even after
tuning for accuracy, the pipeline gives a wrong first answer for about one
column in five, and no numeric setting changes this.

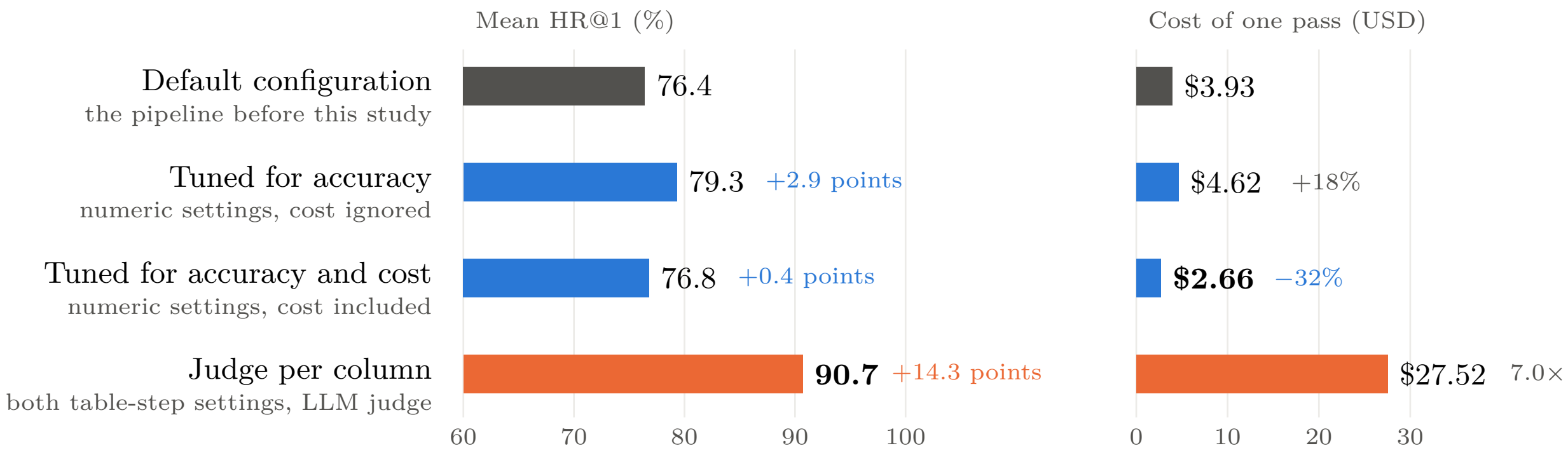
\begin{figure}[t]\centering
\begin{tikzpicture}[font=\footnotesize]
\newcommand{\ax}[1]{(#1-60)*0.115}
\newcommand{\cx}[1]{7.0+(#1)*0.095}
\node[anchor=south west, font=\scriptsize, color=chartmuted] at (0,0.42) {Mean HR@1 (\%)};
\node[anchor=south west, font=\scriptsize, color=chartmuted] at (7.0,0.42) {Cost of one pass (USD)};
\foreach \x in {60,70,80,90,100} {
  \draw[chartgrid, line width=0.5pt] ({\ax{\x}},0.38) -- ({\ax{\x}},-3.45);
  \node[anchor=north, font=\scriptsize, color=chartmuted] at ({\ax{\x}},-3.45) {\x};}
\foreach \x in {0,10,20,30} {
  \draw[chartgrid, line width=0.5pt] ({\cx{\x}},0.38) -- ({\cx{\x}},-3.45);
  \node[anchor=north, font=\scriptsize, color=chartmuted] at ({\cx{\x}},-3.45) {\x};}
\foreach \y/\lab/\sub in {
  0/{Default configuration}/{the pipeline before this study},
  -1/{Tuned for accuracy}/{numeric settings, cost ignored},
  -2/{Tuned for accuracy and cost}/{numeric settings, cost included},
  -3/{Judge per column}/{both table-step settings, LLM judge}} {
  \node[anchor=east] at (-0.2,\y+0.03) {\lab};
  \node[anchor=east, font=\scriptsize, color=chartmuted] at (-0.2,\y-0.31) {\sub};}
\fill[chartmuted] (0,-0.2) rectangle ({\ax{76.4}},0.2);
\fill[chartfront] (0,-1.2) rectangle ({\ax{79.3}},-0.8);
\fill[chartfront] (0,-2.2) rectangle ({\ax{76.8}},-1.8);
\fill[chartchosen] (0,-3.2) rectangle ({\ax{90.7}},-2.8);
\node[anchor=west] at ({\ax{76.4}},0) {76.4};
\node[anchor=west] at ({\ax{79.3}},-1) {79.3};
\node[anchor=west] at ({\ax{76.8}},-2) {76.8};
\node[anchor=west, font=\footnotesize\bfseries] at ({\ax{90.7}},-3) {90.7};
\node[anchor=west, font=\scriptsize, text=chartfront] at ({\ax{79.3}+0.8},-1) {$+2.9$ points};
\node[anchor=west, font=\scriptsize, text=chartfront] at ({\ax{76.8}+0.8},-2) {$+0.4$ points};
\node[anchor=west, font=\scriptsize, text=chartchosen] at ({\ax{90.7}+0.8},-3) {$+14.3$ points};
\fill[chartmuted] ({\cx{0}},-0.2) rectangle ({\cx{3.93}},0.2);
\fill[chartfront] ({\cx{0}},-1.2) rectangle ({\cx{4.62}},-0.8);
\fill[chartfront] ({\cx{0}},-2.2) rectangle ({\cx{2.66}},-1.8);
\fill[chartchosen] ({\cx{0}},-3.2) rectangle ({\cx{27.52}},-2.8);
\node[anchor=west] at ({\cx{3.93}},0) {\$3.93};
\node[anchor=west] at ({\cx{4.62}},-1) {\$4.62};
\node[anchor=west, font=\scriptsize, color=chartmuted] at ({\cx{4.62}+1.05},-1) {$+18\%$};
\node[anchor=west, font=\footnotesize\bfseries] at ({\cx{2.66}},-2) {\$2.66};
\node[anchor=west, font=\scriptsize, text=chartfront] at ({\cx{2.66}+1.05},-2) {$-32\%$};
\node[anchor=west] at ({\cx{27.52}},-3) {\$27.52};
\node[anchor=west, font=\scriptsize, color=chartmuted] at ({\cx{27.52}+1.15},-3) {$7.0\times$};
\end{tikzpicture}
\caption{The configurations of the paper on the same 35 tasks (three seeds;
mean HR@1 over the tasks and cost of one pass over all tasks). The two tuned
configurations were constructed with the group of each task held out. Tuning
the numeric settings gives 2.9 points for 18\% more cost, or the accuracy of
the default for 32\% less cost. Running both settings of the table step and
letting an LLM judge choose per column gives 14.3 points at seven times the
cost. On the larger set of 41 tasks the judge reaches \judgeHR\% at
\judgeRatio{} times the cost of the default (\Cref{tab:judge}). The accuracy
axis starts at 60\%.}
\label{fig:ladder}
\end{figure}

The second part is that the accuracy depends on a choice that numeric tuning
cannot reach. One step of the pipeline, which we call the table step, assigns
each source table to one target table and restricts the search for the columns
of the source table to the assigned table. This choice is of another kind than
a threshold: it decides which answers remain reachable, and an answer that it
excludes cannot be recovered by any later stage or numeric setting. Its effect
is large and it changes sign. The table step raises HR@1 by 30 points on a
MIMIC-III to OMOP task and lowers it by 52 points on a task built from database
views, and on nine of ten test tasks it changes HR@1 by more than the largest
gain of any of six tuned configurations fitted on all tasks. Neither setting is right for all tasks. This part
and the next use a larger set of 41 tasks, the 35 tasks and six that were added
for their structure. On these, mean HR@1 is 75.7\% with the step on, which is
the default, and 83.1\% with it off.

The natural remedy is to choose the setting for each task, and here the
argument turns. Choosing per task is not enough either. Even an oracle that
picks the better setting for each task from the labels stops at 87.8\%, and
the selectors per task that we tested, which do not see the labels, stay
below it. The reason is
that the better setting differs between the columns of one task. In one of our
tasks, two columns of the same source table need opposite settings: a currency
column is mapped correctly only with the step on, because with the step off a
column of the same name in another table is preferred, and an invoice number is
mapped correctly only with the step off, because its correct answer lies
outside the assigned table (\Cref{sec:res-judge}).

The third part follows from this. The choice belongs to the column, and it can
be made after the run, when the answers of both settings are available. The
pipeline runs both settings, and a language model, the judge, chooses among
their candidate answers on the 26\% of columns on which they disagree. This reaches
\judgeHR\% mean HR@1 on the 41 tasks, \judgeOverHind{} points above the
oracle per task, at \judgeRatio{} times the cost of the
default, which is about one cent per column. \Cref{fig:ladder} shows the default,
the tuned configurations and the judge on the 35 tasks on which all of them
were measured; on these the judge reaches 90.7\% at 7.0 times the cost of the
default.

The method behind these results has three parts (\Cref{fig:architecture} in
\Cref{sec:method}). It draws on per-instance algorithm configuration and
algorithm selection, which choose a configuration from features of the
instance~\cite{rice1976algorithmselection,kadioglu2010isac,kerschke2019algorithmselection},
and on multi-objective optimization, which returns a Pareto front instead of
one optimum~\cite{karl2023multiobjective}. Each part answers a difficulty that
these approaches meet in an LLM pipeline, and the three parts are the
contributions of the paper. Every evaluation of a configuration costs model
calls, so we measure each setting of each stage once per labeled task, store
the outputs, and select by replay under cross-validation that holds out whole
groups of related tasks, by accuracy first and cost second and for several
exchange rates between the two
(\Cref{sec:replay,sec:cv,sec:stage-choice,sec:policy-set,sec:res-tuning}). The
labeled tasks number in the tens, too few to fit a selector on task features,
so we measure each value of a structural setting in complete runs and compare
reference points that show whether the setting can be fixed, chosen per task or
only decided per item (\Cref{sec:structural,sec:res-table-step}). And where the
better value depends on the item, a judge chooses per item among the answers of
the competing values; we compare it with a rule on confidences and with a judge
that uses the pipeline's own model (\Cref{sec:judge,sec:res-judge}).

\section{Related work}\label{sec:related}
Choosing an algorithm from features of the instance goes back to
Rice~\cite{rice1976algorithmselection} and is surveyed by Kerschke et
al.~\cite{kerschke2019algorithmselection}. ISAC clusters instances and
configures an algorithm for each cluster~\cite{kadioglu2010isac}, Hydra builds
complementary configurations for a portfolio~\cite{xu2010hydra}, and decision
trees can assign configurations directly~\cite{vilasboas2021decisiontrees}.
Dynamic algorithm configuration goes further and changes settings during a
run~\cite{adriaensen2022dac}. This literature provides the two reference points
that we report throughout: the single best configuration, which is the best
constant, and the virtual best, which is the best configuration for each
instance chosen with the labels. For the table step these are the better of the
two fixed settings and the oracle per task. Our judge moves the
selection from the instance to the item, and the reference points let us state
when that move is necessary.

Multi-objective hyperparameter optimization returns a Pareto front instead of a
single optimum and leaves the choice on the front to the
user~\cite{karl2023multiobjective,deb2002nsga2}. We construct the front from
complete configurations of the pipeline, measured end to end, and report the
whole range, so that a configuration can be chosen for a given budget.

Language models have been used to propose prompts and solutions from past
scores~\cite{yang2023opro}, to guide Bayesian
optimization~\cite{liu2024llambo}, and to compile pipelines of model
calls~\cite{khattab2023dspy}. Cost-aware systems route a query to a cheaper or a
stronger model, or call models in sequence until an answer is
accepted~\cite{chen2023frugalgpt,ong2024routellm}, and Adaptive-RAG selects a
retrieval strategy from the complexity of the
question~\cite{jeong2024adaptiverag}. Language-model judges compare model
outputs~\cite{zheng2023judging}, and self-consistency aggregates several
sampled answers of one model~\cite{wang2022selfconsistency}. Our setting differs from routing in that the choice is among configurations of one pipeline, not among models. Closest to our judge are methods that let a language model select or fuse candidate answers, such as universal self-consistency~\cite{chen2023usc}, LLM-Blender~\cite{jiang2023llmblender} and mixture-of-agents~\cite{wang2024moa}. These combine samples of one model or the outputs of several models. Our judge applies the same kind of selection to the outputs of different configurations of one pipeline, each of which searched a different part of the input, and we use it to answer a configuration question: whether a setting should be chosen per task or per item.

Selecting a configuration and reporting its score on the same data is
optimistic~\cite{cawley2010selection}. In domain generalization, selection on
the test domain is invalid and leave-one-domain-out selection is
valid~\cite{gulrajani2021domaingeneralization}, and algorithm selectors that
appear strong when related instances are split between training and test often
fail on unseen problem classes~\cite{petelin2025pitfalls}. We therefore hold
out sets of related tasks. With few candidates the bias of non-nested
cross-validation is small~\cite{wainer2021nested}. To compare candidates over
held-out sets we use the Bayesian signed-rank test with a region of practical
equivalence~\cite{benavoli2017bayesian}. Hyperparameter importance is known to
differ across datasets~\cite{vanrijn2018importance}; several of our settings
turned out to have none. The last stage of our pipeline decides whether to
answer at all, which is an instance of selective
prediction~\cite{chow1970reject,geifman2017selective}.

Schema matching finds corresponding columns between two schemas, and schema
mapping additionally specifies the computation that fills a
column~\cite{rahm2001survey,dhamankar2004imap,haas2005clio}. Recent systems use
language models with retrieval~\cite{sheetrit2024rematch}, as optimized
language-model programs~\cite{seedat2025matchmaker} or directly on the
schemas~\cite{parciak2024schemamatching}. These systems keep one configuration
for all tasks, although their own results indicate that the best scope differs
by task: ReMatch reports that retrieving two target tables per source attribute
is better than passing the whole target schema on one dataset and worse on
another~\cite{sheetrit2024rematch}.

\section{Problem formulation}\label{sec:problem}
Let $\mathcal{T}$ be a set of tasks. A task $t$ consists of inputs $X_t$ and
labels $Y_t$, where the labels are used only for scoring. A task contains items
$i \in I_t$. In schema matching an item is a source column to be matched, and
from \Cref{sec:application} on we say column. Tasks are related through their
origin, and we describe this relation at two levels. A \emph{family} contains
the tasks built from the same source collection, and a \emph{group} $g(t)$ is
coarser: it merges the families that share a target schema, a source standard
or a generator. The context of a task, $x_t = \phi(X_t)$, consists of
properties that can be computed without labels, such as the sizes of the
schemas, the presence of foreign keys and the coverage of descriptions. A stage
that acts on single items may also use item-level context $x_{t,i}$ that is
available at the moment it acts, for example the confidence of an earlier pass.

The pipeline has stages $s = 1,\dots,S$, and stage $s$ has a finite set of
settings $\Theta_s$. A configuration is a tuple
$\theta = (\theta_1,\dots,\theta_S) \in \Theta = \Theta_1 \times \dots \times \Theta_S$.
Running configuration $\theta$ on task $t$ yields an accuracy $A(\theta,t)$ and
a cost
\begin{equation}
 C(\theta,t) \;=\; \sum_{k \in \mathcal{K}(\theta,t)} \bigl(p^{\rm in}_{m_k}\, n^{\rm in}_k + p^{\rm out}_{m_k}\, n^{\rm out}_k\bigr),
 \label{eq:cost}
\end{equation}
where $\mathcal{K}(\theta,t)$ is the set of model calls made during the run,
$n^{\rm in}_k$ and $n^{\rm out}_k$ are the input and output tokens of call $k$,
and $p^{\rm in}_{m}$ and $p^{\rm out}_{m}$ are the list prices of model $m$.
Both quantities are random because model answers vary between runs, and we
write $A$ and $C$ for their means over repeated runs.

The \emph{default} $\theta^0$ is the configuration that the pipeline had before
this study. A configuration is \emph{fixed} when it is the same tuple for every
task. More generally, the setting of a stage may be given by a rule over the
context, for example a stage that is off for the items on which it cannot act.
We write $\pi$ for a configuration of this general kind and $\pi(x_t)$ for the
tuple that it yields on task $t$, with the item context left implicit, and we
keep the word configuration for both. For a set of tasks $T$ we write
$\bar A(\pi,T) = |T|^{-1}\sum_{t\in T} A(\pi(x_t),t)$ and
$\bar C(\pi,T) = \sum_{t\in T} C(\pi(x_t),t)$, so that accuracy is a mean over
tasks and cost is the total of one pass. Configuration $\pi$ dominates $\pi'$
on $T$ when $\bar A(\pi,T) \ge \bar A(\pi',T)$ and
$\bar C(\pi,T) \le \bar C(\pi',T)$ with at least one strict inequality. How
accuracy is to be traded against cost is the user's decision, which we write as
a goal $u$.

In these terms the paper asks what tuning gains over $\theta^0$, in $\bar A$
and in $\bar C$, on tasks of unseen groups; whether there is a setting for
which no fixed value is adequate; and whether such a setting can be chosen per
task, or an answer has to be chosen for each item from the outputs of several
values. Two reference points separate these cases. For a finite set $\Pi$ of
configurations and tasks $T$, the best fixed choice is
$\max_{\pi\in\Pi}\bar A(\pi,T)$, and the oracle per task, which chooses for each task
with the labels and is the virtual best of the algorithm-selection literature, is
$|T|^{-1}\sum_{t\in T}\max_{\pi\in\Pi} A(\pi(x_t),t)$. No selector that chooses
one member of $\Pi$ per task can exceed the second. Every selection is
evaluated on tasks whose group was not used to make it.

\section{Method}\label{sec:method}

\subsection{Overview}\label{sec:overview}
This section gives the method behind the three parts of the answer: how the
numeric settings are tuned for accuracy and cost
(\Cref{sec:replay,sec:cv,sec:stage-choice,sec:policy-set}), how a setting of
another kind is recognized and measured (\Cref{sec:structural}), and how the
choice is made per item by a judge (\Cref{sec:judge}).

\begin{figure}[t]\centering
\resizebox{\textwidth}{!}{%
\begin{tikzpicture}[font=\scriptsize,
  box/.style={execute at begin node=\hyphenpenalty10000\exhyphenpenalty10000, draw=chartmuted, fill=white, rounded corners=2pt, align=center, inner sep=3pt, text width=2.55cm, minimum height=1.25cm},
  llm/.style={box, fill=chartchosen!14, draw=chartchosen!80!black},
  art/.style={box, fill=chartfront!12, draw=chartfront!80!black},
  dec/.style={draw=chartmuted, fill=white, diamond, aspect=2.1, align=center, inner sep=1pt, text width=1.55cm},
  arr/.style={-{Stealth[length=1.8mm]}, color=chartmuted, line width=0.5pt},
  lab/.style={font=\scriptsize\itshape, color=chartmuted}]
\node[box] (tasks) at (0,0) {Labeled tasks, partitioned into groups};
\node[box] (store) at (3.44,0) {Stored measurements: outputs of every numeric setting, per task and stage};
\node[box] (cv) at (6.88,0) {Selection per stage by replay under grouped cross-validation: accuracy, then cost};
\node[box] (e2e) at (10.31,0) {One configuration per exchange rate $\lambda$, measured end to end};
\node[art] (tuned) at (13.75,0) {Tuned numeric configuration for the chosen $\lambda$};
\draw[arr] (tasks) -- (store); \draw[arr] (store) -- (cv); \draw[arr] (cv) -- (e2e); \draw[arr] (e2e) -- (tuned);
\node[box] (struct) at (3.44,-2.9) {Structural setting with values $\theta^{(1)},\dots,\theta^{(m)}$};
\node[box] (meas) at (6.88,-2.9) {Each value measured in complete runs on every task};
\node[dec] (sign) at (10.31,-2.9) {One value best on all tasks?};
\node[art] (fixed) at (13.75,-2.9) {Fixed value};
\draw[arr] (struct) -- (meas); \draw[arr] (meas) -- (sign);
\draw[arr] (sign) -- node[above, lab] {yes} (fixed);
\node[box] (new) at (0,-5.8) {New task: source and target schemas, no labels};
\node[box] (multi) at (3.44,-5.8) {Pipeline run once with each value $\theta^{(j)}$, numeric settings as tuned};
\node[dec] (agree) at (6.88,-5.8) {First answers agree for the item?};
\node[llm] (judge) at (10.31,-5.8) {\textbf{Judge} (LLM): selects one of the listed candidates};
\node[art] (ans) at (13.75,-5.8) {Answer for every item};
\draw[arr] (new) -- (multi); \draw[arr] (multi) -- (agree);
\draw[arr] (agree) -- node[above, lab] {no} (judge); \draw[arr] (judge) -- (ans);
\draw[arr] (agree.south) -- ++(0,-0.55) -| node[below, lab, pos=0.25] {yes: keep the common answer} (ans.south);
\draw[arr] (sign.south) -- ++(0,-0.5) -| node[above, lab, pos=0.22] {no: decide per item} (multi.north);
\draw[arr] (tuned.south) -- ++(0,-0.7) -| (struct.north);
\begin{scope}[on background layer]
\node[fit=(tasks)(tuned), fill=boxfill, rounded corners=3pt, inner xsep=6pt, inner ysep=9pt] (bg1) {};
\node[fit=(struct)(fixed), fill=boxfill, rounded corners=3pt, inner xsep=6pt, inner ysep=9pt] (bg2) {};
\coordinate (pad) at (0,-6.95);
\node[fit=(new)(ans)(pad), fill=boxfill, rounded corners=3pt, inner xsep=6pt, inner ysep=9pt] (bg3) {};
\end{scope}
\node[anchor=south west, font=\footnotesize\bfseries, color=chartmuted] at (bg1.north west) {Numeric settings: tuned offline for accuracy and cost};
\node[anchor=south west, font=\footnotesize\bfseries, color=chartmuted] at (bg2.north west) {Structural setting: measured, not tuned};
\node[anchor=south west, font=\footnotesize\bfseries, color=chartmuted] at (bg3.north west) {Run time: selection per item};
\end{tikzpicture}}
\caption{Architecture of the method. Top: the numeric settings are tuned
offline; stored outputs are replayed under grouped cross-validation, each stage
is selected by accuracy first and cost second, and each exchange rate $\lambda$
gives one complete configuration. Middle: a structural setting is measured in
complete runs for each of its values; if no value is best on all tasks, it is
not fixed. Bottom: at run time the pipeline is then run once with each value
(in the experiments of this paper with the default numeric settings; the
combination with the tuned settings is not yet measured),
and a judge selects among the candidate answers for every item on which the
first answers differ. The orange box is the only call to a language model
outside the pipeline; blue boxes are the outcomes of each part.}
\label{fig:architecture}
\end{figure}
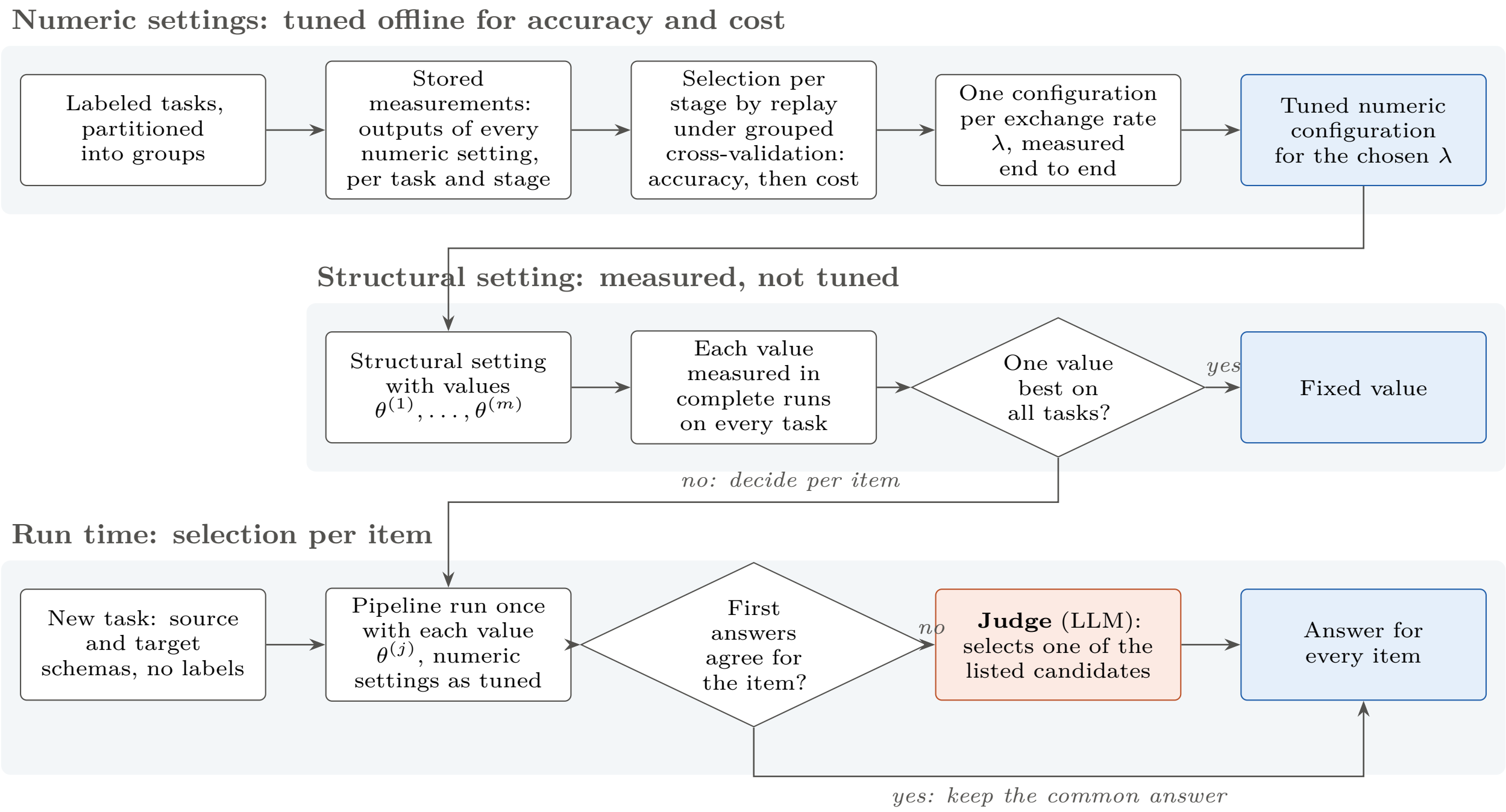

Settings are treated according to how they act (\Cref{fig:architecture}). Most
settings are \emph{numeric}: they change how much the pipeline does, for
example how many candidates it keeps, how much metadata it shows the model, or
whether a single stage such as a verification call runs at all; we count such
switches of single stages among the numeric settings. They are tuned offline
from stored measurements and grouped cross-validation, and the outcome is one
configuration for each exchange rate between accuracy and cost. A few settings
are \emph{structural}: they decide which answers the later stages can reach.
Each value of a structural setting is measured in complete runs, and the
measurements show whether one value can be fixed. Where the better value
changes from one item to the next within a task, the pipeline runs the
competing values and a judge chooses among their answers for each item on
which they disagree.
\Cref{alg:offline,alg:runtime} summarize the offline and the run-time phase.

\subsection{Stored measurements and replay}\label{sec:replay}
Selecting a setting by cross-validation requires the score of every candidate
on every task. If the pipeline were executed for each candidate inside each
fold, a study with $K$ folds, $B$ repetitions and $|\Theta_s|$ settings would
need on the order of $K B |\Theta_s|$ executions per task and stage. We
separate measurement from selection instead. For stage $s$, with the preceding
stages fixed at the settings already selected for them, the pipeline is
executed once for each setting $\theta_s \in \Theta_s$ and each task $t$, and the
outputs of the stage are stored:
\begin{equation}
 O_s(t,\theta_s,r), \qquad r = 1,\dots,R_s ,
\end{equation}
where $r$ indexes repeated executions. From the stored outputs and the labels
we obtain a stage score $a_s(\theta_s,t)$, defined on the output of that stage
alone (for example, whether a correct answer is still among the candidates),
and a stage cost $c_s(\theta_s,t)$, both averaged over the repeats. Scoring any
rule that assigns settings to tasks or items then reduces to a lookup in the
store, and all cross-validation described below is carried out without model
calls.

Settings fall into two classes with different measurement cost. A setting that
only changes how stored outputs are combined, such as a threshold on a
confidence or a weight in a score, is evaluated for all of its values from one
execution. A setting that changes what the model receives, such as the metadata
included in a prompt, requires one execution per value. Stages are measured in
pipeline order, because the outputs of a stage depend on the stages before it.
A stored execution is a single sample of a random outcome, and replaying it in
many folds does not add independent evidence; stages whose outputs vary strongly
between executions are therefore stored with $R_s > 1$. Costs are obtained from
the same executions by \Cref{eq:cost}: calls and candidates are counted, and
tokens are priced at list price without cache discounts, so that the order in
which settings were executed cannot favor any of them.

\subsection{Grouped cross-validation}\label{sec:cv}
The quantity of interest is the effect of a selection rule on a kind of task
that was not available when the rule was chosen. The unit that is held out is
therefore a set of related tasks, a family or a group in the sense of
\Cref{sec:problem}, and never a single task: if one of two tasks that share a
target schema is used for selection and the other for scoring, the score is
optimistic.

Let $\rho_s$ denote a rule for stage $s$, that is, a function from task context
and, where the stage acts on single items, item context to a setting in
$\Theta_s$. A constant setting is the simplest rule. The families are
partitioned into $K$ folds, and the partition is repeated $B$ times with
different random assignments. In each fold the rule is fitted on the families
outside the fold, which we write $\hat\rho_s^{-k}$, and it is scored on each
family $f$ inside the fold as its change against the default,
\begin{equation}
 \Delta_f(\rho_s) \;=\; a_s\bigl(\hat\rho_s^{-k(f)}, f\bigr) - a_s\bigl(\theta^0_s, f\bigr),
 \label{eq:heldout}
\end{equation}
in percentage points, averaged over the $B$ repetitions. The held-out effect of
the rule is the weighted mean of $\Delta_f$ over families with weights
$w_f = \sqrt{n_f}$, where $n_f$ is the number of scored items of the family, so
that large families count more but not in proportion to their size.

Two restrictions determine the data on which a stage is fitted. First, a stage
is fitted only where it can act. Each stage declares the items $I^s_t \subseteq
I_t$ to which it applies; a search over tables linked by foreign keys, for
instance, applies only to columns whose table has such links. A task on which
every setting produces the same output contributes nothing to a comparison of
settings, yet in an average over tasks it reduces every difference, and a
condition that merely separates such tasks improves the fitted score by exactly
zero and is never selected. Excluding them makes the corresponding conditions
part of the stage definition: outside $I^s_t$ the stage is off. Second, a stage
is fitted at the level at which it decides. A stage that makes one decision per
item is fitted on items with their item-level context $x_{t,i}$, which gives
the fit thousands of observations where task averages would give tens, while
entire families remain held out.

To decide whether one rule is better than another we compare the paired
differences $\Delta_f(\rho_s) - \Delta_f(\rho'_s)$ over families with the
Bayesian signed-rank test~\cite{benavoli2017bayesian}. With a region of
practical equivalence of $\pm\varepsilon$ points, the test returns three
posterior probabilities: that $\rho_s$ is worse than $\rho'_s$ by more than
$\varepsilon$, that the two are equivalent, and that $\rho_s$ is better by more
than $\varepsilon$. We set $\varepsilon = 1$ point throughout.

\subsection{Selecting a stage: accuracy first, then cost}\label{sec:stage-choice}
Within a fold, the setting of a stage is selected lexicographically. Let
$\mathcal{C}$ be the candidate rules, and let $a_s(\rho)$ and $c_s(\rho)$ denote
the weighted mean score and the total cost of rule $\rho$ on the training
families. The selected rule is
\begin{equation}
 \rho^\star_s \;=\; \operatorname*{arg\,min}_{\rho \in \mathcal{C}} \; c_s(\rho)
 \quad\text{subject to}\quad a_s(\rho) \;\ge\; \max_{\rho' \in \mathcal{C}} a_s(\rho') - \varepsilon .
 \label{eq:lexi}
\end{equation}
The rule therefore takes the most accurate candidate and then moves to the
cheapest candidate that is practically as accurate. The tolerance $\varepsilon$
is the same region of practical equivalence that the test uses.

The candidates are ordered by complexity: the default setting $\theta^0_s$, the
best constant setting on the training families, and decision trees with one or
two levels of splits over the context features, in which every leaf must
contain at least two families. Trees are grown with the objective
$a_s - \lambda' c_s$ for several values of $\lambda'$, including zero, and the
depth and $\lambda'$ are chosen by an inner cross-validation on the training
families with \Cref{eq:lexi}, so that the outer folds remain untouched by this
choice~\cite{cawley2010selection,wainer2021nested}. A more complex candidate
replaces a simpler one only if the test of \Cref{sec:cv} assigns it a
probability above 0.95 of being better on the held-out families. If two
candidates are rated equivalent, the one that performs less work is retained.
These acceptance rules are fixed before any result is inspected, which prevents
them from being used to justify a preferred outcome. After the comparison, the
selected kind of rule is fitted once more on all families to give the setting
that is deployed.

\subsection{One configuration per exchange rate}\label{sec:policy-set}
\Cref{eq:lexi} yields one configuration, and a single configuration conceals
the trade-off that it resolves. A configuration that gains one point of
accuracy for three times the cost can be Pareto-optimal and still be the wrong
choice for a user with a limited budget. We therefore construct several
configurations that resolve the trade-off differently, measure the accuracy
and cost of each, and leave the choice among them to the user.

Let $\lambda \ge 0$ be an exchange rate between accuracy and cost, defined as
the number of points that a stage must gain to justify doubling its own cost.
For a given $\lambda$, stage $s$ takes the setting
\begin{equation}
 \theta^\lambda_s \;=\; \operatorname*{arg\,max}_{\theta_s \in \Theta_s}\;
 \Delta a_s(\theta_s) - \lambda\, \Delta c_s(\theta_s),
 \qquad
 \Delta c_s(\theta_s) = \frac{c_s(\theta_s)}{c_s(\theta^0_s)} - 1 ,
 \label{eq:lambda}
\end{equation}
where $\Delta a_s$ is the change of the stage score against the default in
points and $\Delta c_s$ is the relative change of the stage cost. Because every
stage is scored on its own measure and its cost is normalized by its own
default, $\lambda$ is applied within a stage and never compares the scores of
two stages. The setting is selected on training families and scored on
held-out families as in \Cref{sec:cv}. The settings of all stages for one
value of $\lambda$ form a complete configuration $\pi^\lambda$, which retains
the conditions that follow from where each stage can act. The value
$\lambda = 0$ ignores cost, and large values give the cheapest configurations. We evaluate
a geometric grid, $\lambda \in \{0, 0.5, 1, 2, 4, 8, 16, 32, 64\}$, and merge
values that lead to identical settings.

Stage scores are proxies for the accuracy of the complete pipeline, and the
costs of stages interact, since a smaller candidate pool shortens every later
prompt. Each configuration $\pi^\lambda$ is therefore executed end to end, and
the set that is reported is defined on the measured results,
\begin{equation}
 \Pi \;=\; \bigl\{\pi^\lambda : \text{no } \pi^{\lambda'} \text{ dominates } \pi^\lambda \text{ on the labeled tasks}\bigr\}.
\end{equation}
The same executions give, for every labeled task, the accuracy and cost of
every configuration.

\begin{algorithm}[t]
\caption{Offline phase: tuning of the numeric settings}\label{alg:offline}
\begin{algorithmic}[1]
\Require labeled tasks $\mathcal{T}$ with families and groups; stages $1,\dots,S$ with settings $\Theta_s$ and default $\theta^0$; exchange rates $\Lambda$
\For{$s = 1,\dots,S$} \Comment{in pipeline order}
  \State execute stage $s$ for every $\theta_s \in \Theta_s$ on every task, with stages $<s$ fixed; store outputs $O_s$
  \State compute $a_s(\theta_s,t)$ and $c_s(\theta_s,t)$ on the items $I^s_t$ where the stage applies
  \For{$\lambda \in \Lambda$}
    \State select $\theta^\lambda_s$ by \Cref{eq:lambda} within grouped cross-validation, by replay of $O_s$
  \EndFor
  \State fix stage $s$ at the setting selected by \Cref{eq:lexi} for the measurement of later stages
\EndFor
\State $\pi^\lambda \gets (\theta^\lambda_1,\dots,\theta^\lambda_S)$ for each $\lambda \in \Lambda$; merge identical configurations
\State execute every $\pi^\lambda$ end to end on $\mathcal{T}$; record $A(\pi^\lambda,t)$ and $C(\pi^\lambda,t)$
\State $\Pi \gets$ configurations not dominated in $(\bar A, \bar C)$
\State \Return $\Pi$
\end{algorithmic}
\end{algorithm}

\subsection{Structural settings}\label{sec:structural}
The procedure above assumes that a setting can be judged by the score of its
stage and that one value, possibly conditioned on context, serves all tasks.
Neither assumption is safe for a class of settings that we call structural. A
setting is \emph{structural} when it determines which answers remain reachable:
an answer that it excludes cannot be recovered by a later stage or by any
numeric setting. A restriction of the search to a part of the input is the
typical case. It removes distractors where the answer lies inside the
restricted part, and it removes the answer where it does not. The definition
says nothing about which of the two effects prevails on a task, or whether the
same one prevails on all tasks; that is measured.

A structural setting is not tuned by a stage score, because its effect appears
only in the final answers. Each of its values $\theta^{(1)},\dots,\theta^{(m)}$
is measured in complete runs on every task, with the numeric settings held
fixed. Three outcomes are possible. If one value is the best on all tasks, it
is fixed like any other setting. If the better value differs between tasks but
is the same for the items of a task, it can be selected per task, by a rule
over context or by a language model that reads results measured on other tasks
(\Cref{app:decider}). If the better value differs between the items of one
task, no selection made before the run can be right for all of them, and the
decision is moved to the item, as described next. Which outcome holds is an
empirical question, and the reference points of \Cref{sec:problem} answer it.
The gap between the best fixed value and the oracle per task
measures what a selection per task could gain. The gap between the
oracle per task and the share of items that at least one value answers
correctly measures what only a selection per item can gain.

\subsection{Selection per item by a judge}\label{sec:judge}
Consider a structural setting whose better value changes from one item to the
next within a task, and for which nothing that is observable before the run
separates the two cases. The best choice per task is then itself a compromise.
The decision is made for each item, and it is made after the run, when the
outputs of the competing values are available as evidence.
Let $\theta^{(1)},\dots,\theta^{(m)}$ be the competing values, with
$\theta^{(1)}$ the default. Running the pipeline with value $j$ returns,
for every item $i$, a ranked list $L^{(j)}_i$ of at most $k$ candidate answers
with confidences, whose first element $y^{(j)}_i$ is the answer of that
setting. The settings disagree on the items
$\mathcal{D}_t = \{i \in I_t : y^{(j)}_i \neq y^{(j')}_i \text{ for some } j, j'\}$.
The final answer is
\begin{equation}
 \hat y_i \;=\;
 \begin{cases}
  y^{(1)}_i & \text{if } i \notin \mathcal{D}_t,\\[2pt]
  J\bigl(i, L^{(\sigma(1))}_i,\dots,L^{(\sigma(m))}_i\bigr) & \text{if } i \in \mathcal{D}_t \text{ and the reply is in } \bigcup_j L^{(j)}_i,\\[2pt]
  y^{(1)}_i & \text{otherwise,}
 \end{cases}
 \label{eq:judge}
\end{equation}
where $J$ is a language model, the judge, and $\sigma$ is a random permutation
drawn for each item. For each setting the judge receives a description of the
part of the input that the setting searched, its ranked candidates with
confidences and descriptions, and the rationale that the pipeline produced for
its first answer. It must return one of the listed candidates; a reply outside
the lists falls back to the default setting.

The settings are presented in permuted order and without their names, so that
the judge cannot acquire a preference for a setting independent of the item. Its
instruction explains why the settings exist and in what respect they differ,
states the objective, and gives no decision rule (\Cref{app:features}). The
judge is a stronger model than the one that runs the pipeline, but its
advantage is not only one of capability: it is the only component that sees the
candidates from every search scope side by side, whereas each pass of the
pipeline saw the candidates of one scope.

Two quantities bound what selection per item can reach. The share of items
for which at least one setting has the correct answer in first position is the
ceiling for any selector that chooses among first answers, and the share for
which a correct answer appears anywhere in the lists is the ceiling for the
judge. Both exceed the oracle per task whenever the
settings are complementary within a task, which is the condition under which
the method is worth its cost. The cost for a task is
\begin{equation}
 C_{\rm col}(t) \;=\; \sum_{j=1}^{m} C\bigl(\theta^{(j)},t\bigr) + |\mathcal{D}_t|\, c_J ,
 \label{eq:colcost}
\end{equation}
with $c_J$ the mean cost of a judge call. Only disputed items reach the judge,
so the second term scales with the disagreement between the settings and not
with the size of the task.

As a baseline that requires no additional model, we use a \emph{confidence
rule} for two settings: the default answer is kept unless the other setting is
more confident by a margin,
$\hat y_i = y^{(2)}_i$ if $\kappa^{(2)}_i > \kappa^{(1)}_i + \gamma$ and
$\hat y_i = y^{(1)}_i$ otherwise, where $\kappa^{(j)}_i$ is the confidence of
the first answer of setting $j$. The margin $\gamma$ is its only fitted
parameter.

\begin{algorithm}[t]
\caption{Run-time phase for a new task $t$}\label{alg:runtime}
\begin{algorithmic}[1]
\Require schemas $X_t$; tuned numeric configuration $\theta^\lambda$; values $\theta^{(1)},\dots,\theta^{(m)}$ of the structural setting, $\theta^{(1)}$ the default
\If{the structural setting has a value that is best on all labeled tasks}
  \State \Return the answers of the pipeline run with $\theta^\lambda$ and that value
\EndIf
\State run the pipeline with $\theta^\lambda$ and each $\theta^{(j)}$, obtaining ranked lists $L^{(j)}_i$ for every item $i$
\For{each item $i$}
  \If{all first answers $y^{(j)}_i$ agree} $\hat y_i \gets y^{(1)}_i$
  \Else\ $\hat y_i \gets$ reply of the judge $J$ on the permuted lists, or $y^{(1)}_i$ if the reply is not listed
  \EndIf
\EndFor
\State \Return $\{\hat y_i\}$
\end{algorithmic}
\end{algorithm}

\section{Application: LLM-based schema matching and mapping}\label{sec:application}
This section describes the pipeline to which the method is applied. Its stages
carry the numeric settings that are tuned in the first part of the results, and
two of its stages form the table step, the structural setting of the second and
third parts.

\begin{figure}[t]\centering
\begin{tikzpicture}[font=\scriptsize,
  box/.style={execute at begin node=\hyphenpenalty10000\exhyphenpenalty10000, draw=chartmuted, fill=white, rounded corners=2pt, align=center, inner sep=3pt, text width=1.82cm, minimum height=1.3cm},
  tstep/.style={box, fill=chartfront!12, draw=chartfront!80!black},
  set/.style={font=\scriptsize\itshape, color=chartmuted, align=center, text width=2.0cm},
  arr/.style={-{Stealth[length=1.8mm]}, color=chartmuted, line width=0.5pt}]
\node[box] (enr) at (0,0) {Name enrichment and embedding};
\node[tstep] (asg) at (2.3,0) {1.\ Table assignment};
\node[box] (plan) at (4.6,0) {2.\ Planning (transformation tasks)};
\node[box] (pool) at (6.9,0) {3.\ Candidate pool};
\node[tstep] (casc) at (9.2,0) {4.\ Cascade to linked tables};
\node[box] (rer) at (11.5,0) {5.\ Reranking and verification};
\node[box] (abst) at (13.8,0) {6.\ Abstention};
\foreach \a/\b in {enr/asg, asg/plan, plan/pool, pool/casc, casc/rer, rer/abst} \draw[arr] (\a) -- (\b);
\node[set] at (2.3,-1.15) {number of calls};
\node[set] at (4.6,-1.25) {proposals, columns shown, metadata};
\node[set] at (6.9,-1.15) {pool cap};
\node[set] at (9.2,-1.15) {threshold $\delta$, bias $b$};
\node[set] at (11.5,-1.25) {metadata shown, verification};
\node[set] at (13.8,-1.15) {threshold $\vartheta$};
\draw[color=chartfront!80!black, line width=0.6pt] (1.35,0.95) -- (1.35,1.1) -- (10.15,1.1) -- (10.15,0.95);
\node[font=\scriptsize, color=chartfront!60!black, anchor=south, align=center, text width=8.6cm] at (5.75,1.1) {table step: \emph{on} restricts the search to the assigned table and its linked tables; \emph{off} retrieves candidates from all tables};
\end{tikzpicture}
\caption{The pipeline for schema matching and mapping. Each source column passes through the
stages from left to right; the settings of each stage are given below it. The
two stages in blue form the table step, which is studied as a structural setting with the values \emph{on} and \emph{off}.}
\label{fig:pipeline}
\end{figure}
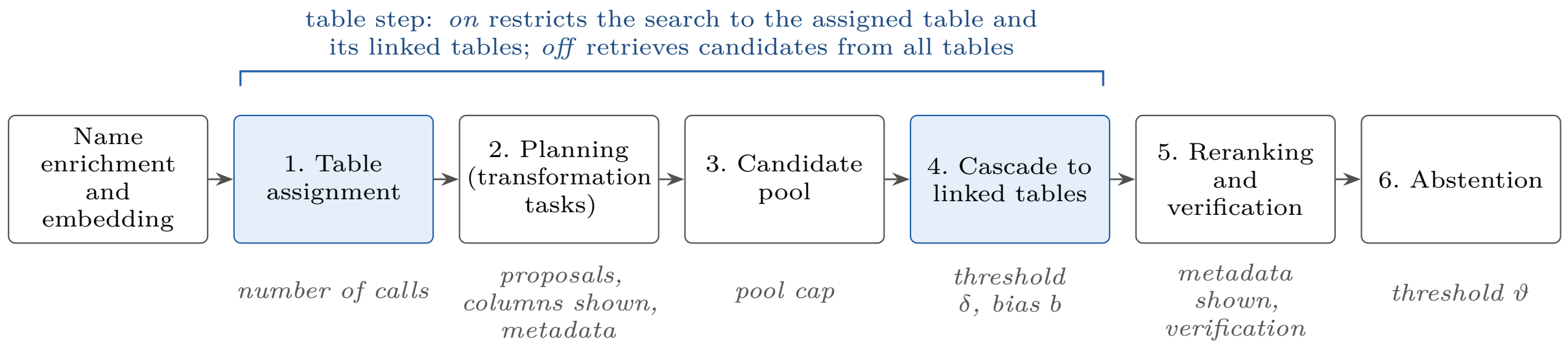

A task consists of a source schema and a target schema. For each source column
the pipeline returns a ranked list of target columns, which is schema matching,
or of transformations over target columns such as the concatenation of two
columns, which is the part of schema mapping that we cover: the pipeline
proposes the rule and its input columns and does not execute it.
The source column is the item of \Cref{sec:problem}. We call a task a
\emph{transformation task} when some of its correct answers are
transformations; in the other tasks every correct answer is a single column.
Every task requires matching, and 13 of the 41 tasks are transformation tasks.
The pipeline is shown in \Cref{fig:pipeline}, and \Cref{tab:stages} lists the
settings of its stages together with the score and the cost on which each stage
is selected.

Before matching, a language model adds alternative names to every column and
all names are embedded; this preparation has no setting in the present study.
The first stage, table assignment, consists of one model call that reads all
source and target tables and assigns to each source table one target table,
which we call its primary table. The call can be repeated and the assignments
combined by majority vote, and the number of calls is the setting of the stage.
On transformation tasks, a planning stage follows in which a model proposes
transformations for each source column from the columns of the primary table
and of the tables linked to it. The third stage builds, for each source column,
a pool of candidate target columns by dense and lexical retrieval, restricted
to the primary table and limited by a cap. The fourth stage, the cascade,
matches the source column inside its primary table first; if the confidence of
that pass is at or below a threshold $\delta$, the tables linked to the primary
table by foreign keys are searched as well, and the two result lists are merged
by confidence after a bias $b$ has been added to the confidences of the
candidates from the primary table. In the fifth stage a model ranks the pool,
optionally followed by a verification call that may exchange the first answer,
and the final confidence $\kappa$ of a candidate combines the score given by
the model with the similarity of the names (\Cref{app:scoring}). The last stage
returns the first answer if $\kappa \ge \vartheta$ and otherwise reports that
there is no match (\Cref{sec:abstention}).

\begin{table}[t]\centering\small
\caption{Stages, their settings, and the score and cost on which each stage is selected.}
\label{tab:stages}
\begin{tabularx}{\linewidth}{@{}L{2.1cm}L{4.4cm}Y@{}}
\toprule Stage & Settings & Score; cost\\\midrule
Table assignment & Number of assignment calls & Share of source tables assigned to an accepted table; assignment calls\\
Planning & Proposals per column, columns shown per table, metadata shown & Share of transformation columns with a complete correct plan; planner tokens\\
Candidate pool & Pool cap, retrieval depth, cosine and BM25 cutoffs & Share of columns whose correct answer is in the pool; pool entries\\
Cascade & Threshold $\delta$, primary bias $b$ & Share of columns whose first result is in an accepted table (routing); matcher calls\\
Reranking & Metadata shown per candidate, verification call, score weights & HR@1; matcher tokens\\
Abstention & Threshold $\vartheta$ & Decision F1; none\\
\midrule
\emph{Table step} & on, off (\Cref{sec:res-table-step}) & HR@1; cost of the run\\
\bottomrule
\end{tabularx}
\end{table}

Table assignment and the cascade together restrict the search for a column to
one target table and the tables linked to it. We refer to this restriction as
the \emph{table step}. It is a structural setting in the sense of
\Cref{sec:structural}, because it decides which target columns remain
reachable, and we study it separately with two settings. With the step
\emph{on}, the pipeline operates as described above. With the step \emph{off},
no table is assigned, and each source column is matched against candidates
retrieved from all target tables. A third setting, which keeps the assignment
as a hint, is reported in \Cref{app:hint}. The table step can help when many
target tables contain columns that resemble each other, such as shared keys and
codes, because fixing the table first excludes the similar columns of other
tables. It can hurt when the columns of one source table originate from many
target tables, because a correct answer outside the assigned table and its
linked tables can no longer be found.

\section{Experimental setup}\label{sec:setup}
The experiments follow the three parts of the answer. The numeric settings are
tuned and measured on 35 tasks. The table step and the judge are measured on
these and six further tasks, 41 in total, with the default numeric settings.
The two sets have different accuracies and costs, and we name the set with
every result.

In every experiment the pipeline runs on Claude Haiku 5.5 with
\code{text-embedding-3-large} embeddings. The judge is Claude Fable 5.2, with Claude Haiku 5.5 as a weaker judge for comparison; the
stronger model is used only for the judge calls. The selection per task of
\Cref{app:decider} uses Claude Opus 5.5 or Claude Fable 5.2.

The study of the numeric settings uses 35 tasks with 2,439 scored columns. The
tasks come from 28 families, which form 12 groups. The decisions per stage
hold out families, and the complete configurations of \Cref{tab:heldout} hold
out groups.
The tasks cover public matching benchmarks (iBench, SchemaNet, RACT, TPC-DI and
entity-integration tasks), clinical data (MIMIC-III and Synthea to the OMOP
common data model, CDISC SDTM to ADaM), geospatial alignments (INSPIRE),
accounts receivable, and seven tasks built from Fivetran's open-source dbt
packages~\cite{fivetran2026dbt}, whose labels are traced from the SQL of the
packages and whose foreign keys are read from the joins that the packages use.
Ten of the 35 tasks (MIMIC-III, Synthea, accounts receivable and the seven
Fivetran tasks) additionally serve as a test set on which six configurations
fitted on all families, the policies P1 to P6, are measured end to end.

The study of the table step adds six tasks, giving 41 tasks with 3,162 scored
columns in 15 groups. The additional tasks were chosen for their structure.
Views of three sample databases (Sakila, Northwind, AdventureWorks) and two dbt
projects (AdventureWorks marts and \code{jaffle_shop}) represent the case in
which one source table draws on several target tables, and an internal
enterprise resource planning (ERP) migration task represents the case in which
many target tables share key columns. Labels for views and dbt models are
derived from their SQL. The AdventureWorks marts task was added first; the
other five were built after the fitted rule of \Cref{sec:res-table-step} had
been written down and before any of them was measured. \Cref{app:tasks} lists
all tasks.

Each task provides correct answers for a subset of its source columns. Only
these columns are scored, and the remaining columns stay in the schemas as
context. A transformation answer is counted as correct when its rule and its
input columns agree with the label. HR@1 is the share of scored source columns
whose first answer is correct. Unless stated otherwise, means are taken over
tasks, so that each task counts equally, and HR@1 is the mean of three
independent runs, which differ because the answers of the model vary between
calls. Differences between configurations are computed before rounding, so that
they can differ by 0.1 from the difference of two rounded values. Cost is the price in US dollars of one pass over the tasks, obtained at
list prices from runs with the response cache disabled; it includes name
enrichment, table assignment, planning and matching. The prices per million
input and output tokens at the time of the runs (November 2026) were \$0.10 and
\$0.50 for Claude Haiku 5.5, \$4 and \$20 for Claude Opus 5.5, and \$8 and \$40
for Claude Fable 5.2. Where a stage is compared with its own default, we report
the relative change in calls or tokens, which does not depend on prices. For
abstention we report the precision, recall and F1 of the decision to answer,
where a column without a correct answer that receives an answer is a false
positive and a column with a correct answer that receives none is a false
negative. For the table step we report both fixed settings and, as a reference,
the \emph{oracle per task}: for each task the setting with the higher HR@1,
chosen with the labels. It is an upper bound for any choice of one setting per
task and not a method that can be run.

\section{Results}\label{sec:results}
The results follow the question of the paper. \Cref{sec:res-tuning} shows what
tuning the numeric settings achieves on unseen groups of tasks,
\Cref{sec:res-table-step} shows where the accuracy is instead,
\Cref{sec:res-judge} shows why the choice has to be made per column and what a
judge reaches, and \Cref{sec:res-summary} states what each gain costs. Results
on the abstention threshold are given in \Cref{app:abstention}.

\FloatBarrier
\subsection{Tuning the numeric settings buys cost and a little accuracy}\label{sec:res-tuning}
On groups of tasks that were held out from tuning, one tuned configuration
gains 2.9 points over the default at 18\% more cost, and another keeps the
accuracy of the default at 32\% lower cost.

To measure complete configurations on unseen groups, we repeated the
construction of \Cref{sec:policy-set} twelve times, each time without one of
the twelve groups and separately for each value of the exchange rate $\lambda$,
and ran the resulting configuration end to end on the tasks of the group that
was left out, with three seeds and the response cache off. This gives one
configuration per value of $\lambda$, from $\lambda = 0$, which ignores cost,
to $\lambda = 64$. From this range we report as \emph{the tuned configuration}
the cheapest one whose mean HR@1 is not below that of the default.
\Cref{tab:heldout} gives the whole range over the 35 tasks.

\begin{table}[htbp]\centering\small
\caption{Configurations constructed without the held-out group and run end to
end on that group (35 tasks, 12 groups, three seeds): cost of one pass over all
tasks and mean HR@1 over the tasks, with changes against the default. The last
column counts the tasks on which the configuration is better and worse than the
default by at least one point. All nine exchange rates are shown, including
those whose configuration is not better than another row on both columns.}
\label{tab:heldout}
\begin{tabular}{@{}lrrrrr@{}}
\toprule Exchange rate & Cost & Cost change & HR@1 & HR@1 change & Better / worse\\\midrule
Default & \$3.93 & --- & 76.4 & --- & ---\\
$\lambda = 0$ & \$4.62 & $+18\%$ & 79.3 & $+2.9$ & 21 / 4\\
$\lambda = 0.5$ & \$4.31 & $+10\%$ & 78.6 & $+2.2$ & 21 / 6\\
$\lambda = 1$ & \$4.28 & $+9\%$ & 78.1 & $+1.7$ & 20 / 7\\
$\lambda = 2$ & \$4.05 & $+3\%$ & 78.1 & $+1.7$ & 21 / 6\\
$\lambda = 4$ & \$3.14 & $-20\%$ & 77.1 & $+0.7$ & 13 / 9\\
$\boldsymbol{\lambda = 8}$ & \textbf{\$2.66} & $\mathbf{-32\%}$ & \textbf{76.8} & $\mathbf{+0.4}$ & 13 / 13\\
$\lambda = 16$ & \$1.79 & $-54\%$ & 75.3 & $-1.1$ & 11 / 12\\
$\lambda = 32$ & \$1.46 & $-63\%$ & 74.2 & $-2.2$ & 9 / 14\\
$\lambda = 64$ & \$1.31 & $-67\%$ & 67.2 & $-9.2$ & 4 / 22\\
\bottomrule
\end{tabular}
\end{table}

The default costs \$3.93 per pass at 76.4\% mean HR@1. Without a penalty on
cost ($\lambda=0$), tuning reaches 79.3\% at \$4.62: a gain of 2.9 points for
18\% more cost, better than the default on 21 tasks and worse on 4. This is the
most that tuning for accuracy obtains. As $\lambda$ grows, cost falls faster
than accuracy. At $\lambda=2$ the gain is still 1.7 points at nearly the cost
of the default, and at $\lambda=8$ the configuration keeps the accuracy of the
default (76.8\% against 76.4\%) at \$2.66, which is 32\% below the cost of the default and
58\% of the cost of the configuration tuned for accuracy alone. This is the
tuned configuration by the rule above. Beyond $\lambda=8$, accuracy is given up
for further savings, down to 67.2\% at one third of the cost. The exchange rate
thus gives the user a range, from somewhat more accuracy at somewhat more cost
to the same accuracy at a third less cost.

The gain in accuracy is not spread evenly. In the same runs, tuning for accuracy gains 7.2 points
on the seven Fivetran tasks and 6.7 points on the eight entity-integration
tasks, and it loses 8.0 points on the two tasks with the OMOP target (MIMIC-III
and Synthea). The settings that it selects search the tables linked to the
assigned table for more columns. They gain where the correct answer often lies
outside the assigned table, and they lose where it lies inside.

Six fixed configurations fitted on all families, which we call the policies P1
to P6 (\Cref{app:stages}), were also measured on ten test tasks and give the same
picture (\Cref{tab:test10}): P2 is within 0.1 points of P1, the configuration
for $\lambda=0$, at 79\% of its cost. The setting selected for each stage,
and the reason for each selection, are given in the same appendix.

The numeric settings are therefore worth tuning: a third of the cost is saved
at no loss of accuracy, or a few points are gained. But the pipeline still
gives a wrong first answer for about one column in five, and the pattern of
gains and losses points to the source. The numeric settings move accuracy by
loosening or tightening one restriction, the assignment of a table, without
being able to remove it or to apply it only where it is right. The largest gain
that any of the six policies reaches on a single test task is 13.7 points
(\Cref{tab:structural}). The next subsection examines the restriction itself,
as one switch, and finds that it moves the accuracy of a task by up to 52
points.

\FloatBarrier
\subsection{The accuracy is in a structural choice: the table step}\label{sec:res-table-step}
Restricting the search for a column to an assigned table raises HR@1 by 30
points on MIMIC-III and lowers it by 52 points on the AdventureWorks views,
and on nine of the ten test tasks its effect is larger than any gain of
numeric tuning.

Except for \Cref{tab:structural}, all runs in this and the next subsection use the 41 tasks and the default
numeric settings, so that the table step (\Cref{sec:structural,sec:application})
is the only thing that varies. With the step on, which is the default, mean
HR@1 on the 41 tasks is 75.7\%; with it off, 83.1\%. The step is better by at
least one point on 10 tasks, worse on 30 and equal on one. The differences are
large in both directions: $+30$ points on MIMIC-III and $+29$ on Synthea, $-52$
on the AdventureWorks views and $-51$ on Fivetran QuickBooks. One small task, an INSPIRE alignment with five scored columns, shows $+33$. That a
restriction of the search can help or hurt follows from its definition; that it
helps on some of our tasks and hurts on others, by this much, is the finding.

\Cref{tab:structural} compares this effect with that of numeric tuning on the
ten test tasks, for which both were measured. For each task it gives the
largest gain and the largest loss against the default over the six policies P1
to P6, and the difference between the table step on and off. The largest gain
that any policy achieves on a task is 13.7 points, and on average the best
policy for a task gains 5.4 points. The table step changes HR@1 by 8.7 to 51.0
points, 22.4 on average, and its effect exceeds the largest numeric gain on
nine of the ten tasks. Numeric settings can lose more than they gain, since the
cheapest policies remove candidates and metadata, but no numeric setting
recovers what the wrong setting of the table step loses. A correct answer that
lies outside the assigned table and its linked tables is not among the
candidates, whatever the later stages do, and a correct answer inside the
assigned table is hidden among similar columns of other tables when the step is
off.

\begin{table}[htbp]\centering\small
\caption{Effect of numeric tuning and of the table step on the ten test tasks,
in points of HR@1 (three seeds). Numeric tuning: largest gain and largest loss
against the default over the policies P1 to P6. Table step: HR@1 with the step
on minus HR@1 with it off. The first three columns come from the runs of the
numeric study and the last from the separate runs of the table-step study, in
which the default scored somewhat differently on some tasks
(\Cref{tab:fivetran}) because the answers of the model vary between runs.}
\label{tab:structural}
\begin{tabular}{@{}lrrrr@{}}
\toprule
 & & \multicolumn{2}{c}{Numeric tuning} & Table step\\
\cmidrule(lr){3-4}
Task & Default HR@1 & Largest gain & Largest loss & on $-$ off\\\midrule
Accounts receivable & 91.8 & $-0.3$ & $-2.4$ & $+16.3$\\
Fivetran Jira & 78.2 & $+10.9$ & $-2.0$ & $-10.2$\\
Fivetran QuickBooks & 25.5 & $+13.7$ & $-5.9$ & $-51.0$\\
Fivetran Stripe & 71.8 & $\phantom{+}0.0$ & $-15.5$ & $-8.7$\\
Fivetran Workday & 65.1 & $+4.1$ & $-4.6$ & $-25.1$\\
Fivetran Xero & 63.7 & $+8.0$ & $-1.0$ & $-21.4$\\
Fivetran Zendesk & 77.0 & $+12.4$ & $-19.9$ & $-19.2$\\
Fivetran Zuora & 78.7 & $+1.9$ & $-1.6$ & $-12.8$\\
MIMIC-III & 66.0 & $+0.4$ & $-13.7$ & $+30.3$\\
Synthea & 90.4 & $+2.6$ & $-7.0$ & $+28.9$\\
\bottomrule
\end{tabular}
\end{table}

Neither fixed setting of the step is therefore right for all tasks, and the
natural next step is to choose the setting for each task. The oracle per task,
which picks the better setting for each task from the labels, gives 87.8\% on the 41 tasks, so a selector that decides
per task before the run can gain at most 4.7 points over \emph{off}, the better
of the two fixed settings. In practice it gains less. A rule over two schema
signals, fitted on 36 tasks and then scored once on five new tasks, was no
better than a constant on the new tasks, and the best of four variants of
selection by a language model reaches 84.9\%, 1.8 points above \emph{off}
(\Cref{app:decider}).

A better selector per task could at most close the gap to 87.8\%. Whether
87.8\% is also the limit of what the table step can give depends on something
that task averages do not show. If the better setting is the same for all
columns of a task, the task is the right unit for the decision. If some columns
are answered correctly only by the setting that is worse for their task as a
whole, the decision belongs to the column. The next subsection asks which of
the two holds.

\FloatBarrier
\subsection{The choice belongs to the column: a judge per column}\label{sec:res-judge}
The better setting of the table step differs between the columns of one task.
A judge that chooses between the answers of the two settings, on the 26\% of
columns on which they differ, reaches \judgeHR\% mean HR@1 on the 41 tasks,
above the oracle per task, at \judgeRatio{} times the cost
of the default.

Two columns of the Fivetran Zuora task show the problem (\Cref{tab:example}).
Both belong to one source table, which the table step assigns to the target
table \code{invoice_item}. For \code{home_currency}, described as the home
currency in which the customer is billed, the step on gives the correct answer.
The step off prefers a column of the same name in another table, with a
slightly higher confidence (0.726 against 0.699). For \code{invoice_number},
described as the unique identification number of the invoice, the correct
answer lies outside the assigned table, so the step on cannot give it and the
step off does. No setting chosen for this task, or even for this source table,
is right for both columns. The judge chose the correct answer in both cases. In
this task the first answers of the two settings differ on 22 columns; the judge
chose correctly on 4 columns for which only \emph{on} was right and on 16 for
which only \emph{off} was right.

\begin{table}[htbp]\centering\small
\caption{Two columns of one source table of the Fivetran Zuora task,
\texttt{zuora\_\_line\_item\_history}, which the table step assigns to
\texttt{invoice\_item}. Each needs a different setting of the table step.}
\label{tab:example}
\begin{tabular}{@{}llll@{}}
\toprule
Source column & Answer with the step on & Answer with the step off & Judge chose\\\midrule
\code{home_currency} & \code{INVOICE_ITEM.HOME_CURRENCY} & \code{RATE_PLAN_CHARGE.HOME_CURRENCY} & answer of \emph{on}\\
 & correct & wrong & correct\\[3pt]
\code{invoice_number} & \code{INVOICE_ITEM.INVOICE_ID} & \code{INVOICE.INVOICE_NUMBER} & answer of \emph{off}\\
 & wrong & correct & correct\\
\bottomrule
\end{tabular}
\end{table}

To measure how often this happens, we apply selection per column
(\Cref{sec:judge}) to all 41 tasks. Every source column whose table has an
assignment is matched with the step on and with it off, both on Claude Haiku
5.5, and where the two first answers differ, a judge chooses one target column
from the two candidate lists.

\Cref{tab:replay} compares the ways of combining the two settings on equal
terms. The candidates of both settings were saved in one run, and every method
was applied to the same saved candidates, so that the differences between the
rows are not affected by run-to-run variation. The first answers differ on 816
of the 3,162 columns, and only these reach a judge. Because the table comes
from one run, its values differ slightly from the means over three seeds
reported for the same settings elsewhere.

\begin{table}[htbp]\centering\small
\caption{Combining the two settings of the table step on the same saved
candidates (one run, 41 tasks, 3,162 columns). The 8 tasks are MIMIC-III and
the seven Fivetran tasks, on which the margin of the confidence rule was
chosen. The last column counts the tasks on which the method is within one
point of, or above, the better of the two settings for that task. The cost of
a judge is the total of its 816 calls. In the control, the judge sees the
candidates of one setting on the same 816 columns.}
\label{tab:replay}
\begin{tabular}{@{}lrrrrr@{}}
\toprule
 & \multicolumn{3}{c}{Mean HR@1} & Judge & Within 1 point of\\
\cmidrule(lr){2-4}
Method & 33 tasks & 8 tasks & All 41 & cost & the better setting\\\midrule
Table step on (default) & 79.5 & 67.7 & 77.2 & --- & 13/41\\
Table step off & 83.7 & 79.9 & 82.9 & --- & 30/41\\
Confidence rule & 88.7 & 83.4 & 87.7 & --- & 25/41\\
Judge: Claude Haiku 5.5 & 88.4 & 85.0 & 87.8 & \$0.33 & 32/41\\
Oracle per task & 89.9 & 83.4 & 88.6 & --- & 41/41\\
\textbf{Judge: Claude Fable 5.2} & \textbf{92.1} & \textbf{88.5} & \textbf{91.4} & \$21.5 & \textbf{39/41}\\
\midrule
\multicolumn{6}{@{}l}{\emph{Control: Claude Fable 5.2 sees the candidates of one setting only}}\\
Candidates of \emph{on} only & 81.0 & 67.8 & 78.4 & \$19.8 & 14/41\\
Candidates of \emph{off} only & 90.3 & 86.3 & 89.5 & \$16.5 & 37/41\\
\midrule
A correct first answer in at least one setting & 92.8 & 88.5 & 92.0 & & \\
\bottomrule
\end{tabular}
\end{table}

The table supports three statements. First, the decision belongs to the
column. A correct first answer exists in at least one of the two settings for
92.0\% of the columns of a task on average, whereas the better setting for each task reaches 88.6\%.
If the better setting were the same for all columns of a task, the two values
would be equal; the difference of 3.4 points is accuracy that no choice per
task can obtain, however it is made. Second, the judge with Claude Fable 5.2
obtains most of it. It reaches 91.4\%. Measured on the unrounded values, this
is 8.4 points above \emph{off}, the better of the two fixed settings (95\%
bootstrap interval over tasks: 5.3 to 12.2), and 2.7 points above the oracle
per task (0.9 to 4.4). Third, the strength of the judge
matters. The confidence rule, which needs no further model, and the judge with
Claude Haiku 5.5, the model that runs the pipeline, both reach about 87.7\%:
they are about 4.7 and 4.8 points above \emph{off}, and neither exceeds the
oracle per task. The stronger judge adds 3.6 points to the
weaker one (2.1 to 5.4). Every reply of both judges named a listed candidate.

The two control rows are a control that separates the judge from
the settings it compares. On the same 816 columns, the same model was shown the
five candidates of one setting only and asked to choose among them. With the
candidates of \emph{on} it reaches 78.4\%, little above \emph{on} itself: an
answer outside the assigned table is not among the candidates, and in 65 of the
816 calls the judge replied that none of the listed candidates fits. This is
the statement of \Cref{sec:res-table-step} seen from the other side, since a
stronger model cannot recover what the table step has excluded. With the
candidates of \emph{off} the judge reaches 89.5\%, so a large part of the gain
comes from a second, stronger reading of the columns on which the two settings
disagree. With the candidates of both settings it reaches 91.4\%, which is 1.8
points more (95\% interval $-0.1$ to 4.3), better on 16 tasks and worse on 2.
The complementarity of the two settings is not an effect of running the
pipeline twice. For 19 of the tasks a second run of both settings was saved. On
these, a correct first answer exists in \emph{on} or \emph{off} for 88.8\% of
the columns, against 80.3\% in two runs of \emph{off} and 76.3\% in two runs
of \emph{on}. The two settings thus serve the judge twice: their disagreement identifies the
26\% of columns that are worth a call to the stronger model, and the answers of
the restricted search add candidates that the search over all tables ranks too
low.

\Cref{tab:judge} gives complete runs with three seeds, which also give the
cost. The margin of the confidence rule (0.10) was chosen on the eight tasks
named above; the judge has no fitted number.

\begin{table}[htbp]\centering\small
\caption{The table step on the 41 tasks in complete runs: mean HR@1 over three
seeds and cost of one pass over all tasks. The 33 tasks of the first column
were not used to choose the margin of the confidence rule; the 8 tasks of the
second were. The oracle per task costs one pass, but it can be chosen
only with the labels.}
\label{tab:judge}
\begin{tabular}{@{}lrrrr@{}}
\toprule
 & \multicolumn{3}{c}{Mean HR@1} & Cost\\
\cmidrule(lr){2-4}
Method & 33 tasks & 8 tasks & All 41 & (all 41)\\\midrule
Table step on (default) & 78.4 & 64.9 & 75.7 & \$5.79\\
Table step off & 83.9 & 79.6 & 83.1 & \$6.15\\
Oracle per task & 88.8 & 83.4 & 87.8 & \$5.85\\
Confidence rule & 88.1 & 84.4 & 87.4 & \$11.64\\
\textbf{Judge: Claude Fable 5.2} & \textbf{\judgeHRa} & \textbf{\judgeHRb} & \textbf{\judgeHR} & \judgeCost\\
\bottomrule
\end{tabular}
\end{table}

In the complete runs the judge reaches \judgeHR\% mean HR@1 over the 41 tasks,
\judgeOverOn{} points above the default, \judgeOverOff{} points above
\emph{off} and \judgeOverHind{} points above the oracle per task (2.7 points in the
comparison on one saved run of \Cref{tab:replay}). It is within one point of, or above, the better of the two
settings on \judgeWithin{} of the 41 tasks. One run with the judge costs \judgeCost,
\judgeRatio{} times the cost of the default: two runs of the pipeline on the
small model and the calls to the judge, at about \$0.03 per call. The
confidence rule costs \$11.64, a little less than the two single passes
together (\$11.94), because the two passes run in one process and share name
enrichment and table assignment.

\Cref{tab:judge-groups} shows that the judge follows the better setting,
whichever it is. In the groups in which the table step helps (OMOP, CDISC,
accounts receivable, internal ERP), it stays at or near the level of \emph{on};
in the groups in which the step hurts (Fivetran, views and marts, entity
integration), it reaches or exceeds the level of \emph{off}; and where neither
setting is good (INSPIRE), it is above both.

\begin{table}[htbp]\centering\small
\caption{Mean HR@1 per group for the table step on the 41 tasks in complete runs (three seeds).
Bold: the best method of the row.}
\label{tab:judge-groups}
\begin{tabular}{@{}lrrrrr@{}}
\toprule Group & Tasks & On & Off & Confidence rule & Judge\\\midrule
Fivetran & 7 & 64.7 & 85.9 & 86.8 & \textbf{92.1}\\
Entity integration (SINT) & 8 & 81.2 & 93.8 & 90.7 & \textbf{95.3}\\
RACT / Sakila & 8 & 78.2 & 83.7 & 85.9 & \textbf{89.1}\\
INSPIRE & 3 & 59.3 & 42.6 & 57.5 & \textbf{70.4}\\
OMOP (MIMIC-III, Synthea) & 2 & 78.2 & 48.6 & \textbf{82.0} & 81.5\\
AdventureWorks (marts, views) & 2 & 44.0 & 90.5 & 89.9 & \textbf{93.5}\\
Northwind & 2 & 73.3 & 95.1 & 93.0 & \textbf{97.6}\\
iBench & 2 & 76.3 & 83.2 & 92.5 & \textbf{95.9}\\
Accounts receivable & 1 & \textbf{91.8} & 75.5 & 89.8 & 90.1\\
CDISC & 1 & 95.0 & 87.9 & 94.4 & \textbf{95.5}\\
Internal ERP & 1 & 99.0 & 89.9 & 98.2 & \textbf{99.1}\\
TPC-DI & 1 & 95.9 & 97.3 & \textbf{99.0} & \textbf{99.0}\\
SchemaNet Bank & 1 & 88.2 & 90.2 & 92.2 & \textbf{98.0}\\
Real estate & 1 & 85.7 & \textbf{100.0} & \textbf{100.0} & \textbf{100.0}\\
\code{jaffle_shop} & 1 & \textbf{100.0} & 81.0 & \textbf{100.0} & \textbf{100.0}\\
\bottomrule
\end{tabular}

\end{table}

The seven Fivetran tasks, from which the example above is taken, show the same
effect on the scale of a group (\Cref{tab:fivetran}). Their source tables are
analytics models that combine columns of several raw tables, so that the table
step is harmful on average: \emph{on} reaches 64.7\% and \emph{off} 85.9\%. Yet
some columns of each model do come from one main table, and for these the
restricted search is the more reliable one. The judge reaches
\judgeFivetran\%, which is \judgeFivetranOverOn{} points above the default and
\judgeFivetranOverOff{} points above \emph{off}.

\begin{table}[htbp]\centering\small
\caption{The seven Fivetran tasks: HR@1 in complete runs (mean of three seeds).}
\label{tab:fivetran}
\begin{tabular}{@{}lrrrr@{}}
\toprule Task & Scored columns & On & Off & Judge\\\midrule
Jira & 49 & 80.3 & 90.5 & \textbf{98.0}\\
QuickBooks & 17 & 31.4 & 82.3 & \textbf{84.3}\\
Stripe & 84 & 67.9 & 76.6 & \textbf{83.3}\\
Workday & 65 & 55.9 & 81.0 & \textbf{95.9}\\
Xero & 67 & 63.7 & 85.1 & \textbf{91.5}\\
Zendesk & 97 & 76.0 & \textbf{95.2} & 94.2\\
Zuora & 86 & 77.9 & 90.7 & \textbf{97.7}\\
\midrule
Mean & & 64.7 & 85.9 & \textbf{92.1}\\
\bottomrule
\end{tabular}

\end{table}

The judge thus gives the highest accuracy of all configurations, and it is also
the most expensive one. What remains is a practical question: what does each
gain in accuracy cost, and when is the judge worth it?

\subsection{Summary: what each gain costs}\label{sec:res-summary}
\begin{figure}[htbp]\centering
\begin{tikzpicture}
\begin{groupplot}[group style={group size=2 by 1, horizontal sep=1.7cm}, width=0.5\textwidth, height=0.4\textwidth,
  tick label style={font=\scriptsize, color=chartmuted}, label style={font=\scriptsize}, axis lines=left,
  axis line style={chartmuted}, grid=major, grid style={chartgrid}, clip=false,
  every axis title/.style={at={(0,1.02)}, anchor=south west, font=\small}]
\nextgroupplot[title={Numeric settings, 35 tasks, groups held out}, xlabel={Cost of one pass (USD)}, ylabel={Mean HR@1 (\%)}, xmin=1, xmax=5, ymin=66, ymax=81]
\addplot[thick, color=chartfront] coordinates {(1.31,67.2) (1.46,74.2) (1.79,75.3) (2.66,76.8) (3.14,77.1) (4.05,78.1) (4.28,78.1) (4.31,78.6) (4.62,79.3)};
\addplot[only marks, mark=*, mark size=1.8pt, color=chartmuted] coordinates {(1.31,67.2) (1.46,74.2) (1.79,75.3) (3.14,77.1) (4.05,78.1) (4.28,78.1) (4.31,78.6) (4.62,79.3)};
\addplot[dashed, color=chartmuted] coordinates {(1,76.4) (5,76.4)};
\addplot[only marks, mark=square*, mark size=2pt, color=chartother] coordinates {(3.93,76.4)};
\addplot[only marks, mark=*, mark size=2.6pt, color=chartchosen] coordinates {(2.66,76.8)};
\node[font=\scriptsize, anchor=north west] at (axis cs:3.95,76.2) {default};
\node[font=\scriptsize, anchor=north, text=chartchosen] at (axis cs:2.95,75.9) {tuned ($\lambda=8$)};
\node[font=\scriptsize, anchor=south] at (axis cs:4.62,79.6) {$\lambda=0$};
\node[font=\scriptsize, anchor=west] at (axis cs:1.37,67.1) {$\lambda=64$};
\nextgroupplot[title={Table step, 41 tasks}, xlabel={Cost of one pass (USD, log scale)}, xmode=log, log ticks with fixed point,
  xtick={5,10,20,40}, xmin=4.6, xmax=60, ymin=74, ymax=93]
\addplot[thick, color=chartfront] coordinates {(6.15,83.08) (11.64,87.36) (\judgeCostNum,\judgeHR)};
\addplot[dashed, color=chartmuted] coordinates {(4.6,87.76) (60,87.76)};
\addplot[only marks, mark=square*, mark size=1.8pt, color=chartother] coordinates {(5.79,75.74)};
\addplot[only marks, mark=*, mark size=1.8pt, color=chartmuted] coordinates {(6.15,83.08) (11.64,87.36)};
\addplot[only marks, mark=*, mark size=2.6pt, color=chartchosen] coordinates {(\judgeCostNum,\judgeHR)};
\node[font=\scriptsize, anchor=west] at (axis cs:6.1,75.6) {on (default)};
\node[font=\scriptsize, anchor=north west] at (axis cs:6.3,82.9) {off};
\node[font=\scriptsize, anchor=south west, color=chartmuted] at (axis cs:4.8,87.9) {oracle per task};
\node[font=\scriptsize, anchor=north west] at (axis cs:11.4,87.0) {confidence rule};
\node[font=\scriptsize, anchor=south east, text=chartchosen] at (axis cs:31,91.3) {judge};
\end{groupplot}
\end{tikzpicture}
\caption{Accuracy against cost in complete runs. Left: the configurations of
\Cref{tab:heldout}, one for each exchange rate $\lambda$, on held-out groups;
the dashed line is the accuracy of the default. Right: the table step on the 41 tasks with
the default numeric settings; the line joins the fixed setting \emph{off}, the
confidence rule and the judge, and the dashed line is the better of \emph{on}
and \emph{off} chosen for each task with the labels (oracle per task).}
\label{fig:fronts}
\end{figure}
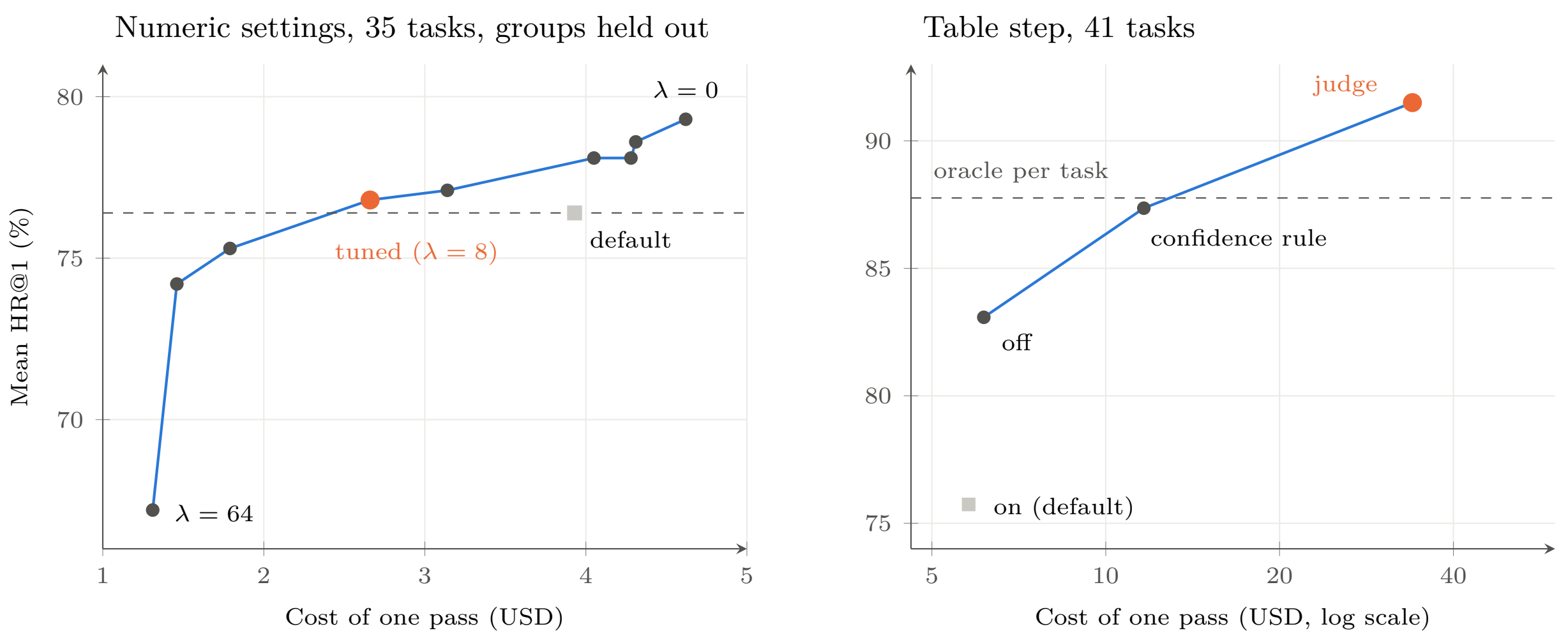

\Cref{tab:summary} places the configurations of the three parts side by side,
and \Cref{fig:fronts} shows them as accuracy against cost. Tuning the numeric
settings gives up to 2.9 points, or the accuracy of the default at about a
third less cost.
The two fixed settings of the table step differ by 7.4 points on average and by
up to 52 points on one task. Selection per column gives more accuracy than
either setting, or the better of the two per task, at a multiple of the cost.

For a practitioner, the results on the 41 tasks are a choice among three levels
of cost. Turning the table step off gives 83.1\% at 1.06 times the cost of the
default. The confidence rule gives 87.4\% at 2.0 times the cost and needs no
stronger model. The judge gives \judgeHR\% at \judgeRatio{} times the cost. One
run with the judge over the 3,162 scored columns costs \judgeCost, about one
cent per column; a call to the judge costs about \$0.03 and is made for 26\% of
the columns. What matters for the choice is the cost of each additional correct
answer. Against the step off, the confidence rule answers about 120 more
columns correctly for \$5.49 more, about 5 cents per additional correct column,
and the judge about 217 more for \$27.82 more, about 13 cents each. Going from
the confidence rule to the judge costs about 23 cents per additional correct
column. The judge is therefore the right choice where a wrong mapping costs
more than about 25 cents to find and repair, and the confidence rule where it
costs more than about 5 cents. The tuned numeric settings are independent of
this choice and are expected to lower the cost of each run.

\begin{table}[htbp]\centering\small
\caption{The configurations of the three parts. Cost is the cost of one pass
over the tasks named in the second column; the two sets of tasks have different
costs and accuracies and are not comparable with each other.}
\label{tab:summary}
\begin{tabular}{@{}llrrl@{}}
\toprule
Configuration & Tasks and protocol & HR@1 & Cost & Against the reference\\\midrule
\multicolumn{5}{@{}l}{\emph{Numeric settings}}\\
Default & 35 tasks, three seeds & 76.4 & \$3.93 & reference\\
Tuned for accuracy ($\lambda=0$) & 35 tasks, groups held out & 79.3 & \$4.62 & $+2.9$ points, $+18\%$ cost\\
Tuned with cost ($\lambda=8$) & 35 tasks, groups held out & 76.8 & \$2.66 & $+0.4$ points, $-32\%$ cost\\
\midrule
\multicolumn{5}{@{}l}{\emph{Table step, default numeric settings}}\\
On (default) & 41 tasks, three seeds & 75.7 & \$5.79 & reference\\
Off & 41 tasks, three seeds & 83.1 & \$6.15 & $+7.4$ points, $+6\%$ cost\\
Oracle per task & 41 tasks, three seeds & 87.8 & \$5.85 & $+12.0$ points; needs the labels\\
\midrule
\multicolumn{5}{@{}l}{\emph{Selection per column between on and off}}\\
Confidence rule & 41 tasks, three seeds & 87.4 & \$11.64 & $+11.6$ points, $2.0\times$ cost\\
Judge: Claude Fable 5.2 & 41 tasks, three seeds & \judgeHR & \judgeCost & $+\judgeOverOn$ points, $\judgeRatio\times$ cost\\
\bottomrule
\end{tabular}
\end{table}

\FloatBarrier

\section{Conclusion and future work}\label{sec:conclusion}
We tuned every numeric setting of an LLM pipeline for schema matching and mapping, for
accuracy and cost together, and asked what this buys and where the remaining
accuracy is. Tuning the numeric settings buys cost and a little accuracy: on
35 tasks, with groups of tasks held out from tuning, one tuned configuration
gains 2.9 points over the default (79.3\% against 76.4\% HR@1) at 18\% more
cost, and another keeps the accuracy of the default at 32\% lower
cost. The accuracy depends on a structural choice that numeric tuning cannot
reach. Restricting the search for a column to one assigned table changes HR@1
by $+30$ points on one task and by $-52$ points on another, so that neither
setting is right for all tasks: on 41 tasks, mean HR@1 is 75.7\% with the
restriction and 83.1\% without it. Choosing the setting per task is not enough
either, because the better setting differs between the columns of one task;
an oracle that picks the better setting for each task reaches 87.8\%. When the pipeline runs both settings and an
LLM judge chooses among their candidate answers on the columns on which they disagree,
mean HR@1 on the 41 tasks reaches \judgeHR\%, at \judgeRatio{} times the cost of
the default. That is about one cent per column processed and about 13 cents
for each additional correct answer over the better of the two fixed settings,
so the judge is worth its cost where a wrong mapping is more expensive than
that to find and repair. The tuned numeric settings are expected to lower the
cost of each run.

The method is not specific to schema matching. Stored measurements make grouped
cross-validation free of model calls, and an accuracy-then-cost rule with an
exchange rate yields configurations for different budgets. The best fixed
value, the oracle per task and the share of items
that at least one value answers correctly show at which level a setting has to
be decided, and a judge makes the decision per item where that is required.

Several next steps follow from the results. The numeric settings and the table
step were varied separately, and the tuned numeric configuration remains to be
measured together with selection per column, where it should lower the cost of
each of the two passes. The cost of the judge can be reduced: the confidence
rule and a judge with the pipeline's own model reach the same accuracy in our
replay, so a stronger judge that is called only for the columns on which the
confidence rule is uncertain is the first candidate. Selection of a
configuration per task by a language model was effective only when a task of
the same group had been measured (\Cref{app:decider}), and a larger pool of
measured tasks would show how its value grows with the coverage of the
evidence. The prompt of the reranker can be treated as a setting, searched
within each fold from the errors on the training groups. And the study can be
repeated with other models in the pipeline and with selection per item applied
to further structural settings.

\bibliographystyle{unsrtnat}
\bibliography{references}

@article{kerschke2019algorithmselection,
  author  = {Kerschke, Pascal and Hoos, Holger H. and Neumann, Frank and Trautmann, Heike},
  title   = {Automated Algorithm Selection: Survey and Perspectives},
  journal = {Evolutionary Computation},
  year    = {2019},
  volume  = {27},
  number  = {1},
  pages   = {3--45},
  doi     = {10.1162/evco_a_00242},
  url     = {https://arxiv.org/abs/1811.11597}
}

@inproceedings{kadioglu2010isac,
  author    = {Kadioglu, Serdar and Malitsky, Yuri and Sellmann, Meinolf and Tierney, Kevin},
  title     = {{ISAC}---Instance-Specific Algorithm Configuration},
  booktitle = {ECAI 2010},
  series    = {Frontiers in Artificial Intelligence and Applications},
  volume    = {215},
  publisher = {IOS Press},
  year      = {2010},
  pages     = {751--756},
  doi       = {10.3233/978-1-60750-606-5-751},
  url       = {https://ebooks.iospress.nl/doi/10.3233/978-1-60750-606-5-751}
}

@inproceedings{xu2010hydra,
  author    = {Xu, Lin and Hoos, Holger H. and Leyton-Brown, Kevin},
  title     = {{Hydra}: Automatically Configuring Algorithms for Portfolio-Based Selection},
  booktitle = {Proceedings of the Twenty-Fourth AAAI Conference on Artificial Intelligence},
  publisher = {AAAI Press},
  year      = {2010},
  pages     = {210--216},
  doi       = {10.1609/aaai.v24i1.7565},
  url       = {https://ojs.aaai.org/index.php/AAAI/article/view/7565}
}

@article{vilasboas2021decisiontrees,
  author  = {Vilas Boas, Matheus Guedes and Santos, Haroldo Gambini and Merschmann, Luiz Henrique de Campos and Vanden Berghe, Greet},
  title   = {Optimal Decision Trees for the Algorithm Selection Problem: Integer Programming Based Approaches},
  journal = {International Transactions in Operational Research},
  year    = {2021},
  volume  = {28},
  number  = {5},
  pages   = {2759--2781},
  doi     = {10.1111/itor.12724},
  url     = {https://arxiv.org/abs/1907.02211}
}

@article{adriaensen2022dac,
  author  = {Adriaensen, Steven and Biedenkapp, Andr{\'e} and Shala, Gresa and Awad, Noor and Eimer, Theresa and Lindauer, Marius and Hutter, Frank},
  title   = {Automated Dynamic Algorithm Configuration},
  journal = {Journal of Artificial Intelligence Research},
  year    = {2022},
  volume  = {75},
  pages   = {1633--1699},
  doi     = {10.1613/jair.1.13922},
  url     = {https://jair.org/index.php/jair/article/view/13922}
}

@incollection{rice1976algorithmselection,
  author    = {Rice, John R.},
  title     = {The Algorithm Selection Problem},
  booktitle = {Advances in Computers},
  volume    = {15},
  publisher = {Elsevier},
  year      = {1976},
  pages     = {65--118},
  doi       = {10.1016/S0065-2458(08)60520-3},
  url       = {https://www.sciencedirect.com/science/article/pii/S0065245808605203}
}

@article{rahm2001survey,
  author  = {Rahm, Erhard and Bernstein, Philip A.},
  title   = {A Survey of Approaches to Automatic Schema Matching},
  journal = {The VLDB Journal},
  year    = {2001},
  volume  = {10},
  pages   = {334--350},
  doi     = {10.1007/s007780100057},
  url     = {https://link.springer.com/article/10.1007/s007780100057}
}

@inproceedings{dhamankar2004imap,
  author    = {Dhamankar, Robin and Lee, Yoonkyong and Doan, AnHai and Halevy, Alon and Domingos, Pedro},
  title     = {{iMAP}: Discovering Complex Semantic Matches between Database Schemas},
  booktitle = {Proceedings of the 2004 ACM SIGMOD International Conference on Management of Data},
  year      = {2004},
  pages     = {383--394},
  publisher = {ACM},
  doi       = {10.1145/1007568.1007612},
  url       = {https://homes.cs.washington.edu/~pedrod/papers/sigmod04.pdf}
}

@inproceedings{haas2005clio,
  author    = {Haas, Laura M. and Hern{\'a}ndez, Mauricio A. and Ho, Howard and Popa, Lucian and Roth, Mary},
  title     = {{Clio} Grows Up: From Research Prototype to Industrial Tool},
  booktitle = {Proceedings of the 2005 ACM SIGMOD International Conference on Management of Data},
  year      = {2005},
  pages     = {805--810},
  publisher = {ACM},
  doi       = {10.1145/1066157.1066252},
  url       = {https://research.ibm.com/publications/clio-grows-up-from-research-prototype-to-industrial-tool}
}

@misc{sheetrit2024rematch,
  author        = {Sheetrit, Eitam and Brief, Menachem and Mishaeli, Moshik and Elisha, Oren},
  title         = {{ReMatch}: Retrieval Enhanced Schema Matching with {LLMs}},
  year          = {2024},
  eprint        = {2403.01567},
  archivePrefix = {arXiv},
  primaryClass  = {cs.DB},
  doi           = {10.48550/arXiv.2403.01567},
  url           = {https://arxiv.org/abs/2403.01567},
  note          = {Version 2, 30 May 2024}
}

@inproceedings{seedat2025matchmaker,
  author    = {Seedat, Nabeel and van der Schaar, Mihaela},
  title     = {Bootstrapping Self-Improvement of Language Model Programs for Zero-Shot Schema Matching},
  booktitle = {Proceedings of the 42nd International Conference on Machine Learning},
  year      = {2025},
  volume    = {267},
  series    = {Proceedings of Machine Learning Research},
  pages     = {53791--53826},
  publisher = {PMLR},
  url       = {https://proceedings.mlr.press/v267/seedat25a.html}
}

@inproceedings{jeong2024adaptiverag,
  author    = {Jeong, Soyeong and Baek, Jinheon and Cho, Sukmin and Hwang, Sung Ju and Park, Jong C.},
  title     = {{Adaptive-RAG}: Learning to Adapt Retrieval-Augmented Large Language Models through Question Complexity},
  booktitle = {Proceedings of the 2024 Conference of the North American Chapter of the Association for Computational Linguistics: Human Language Technologies (Volume 1: Long Papers)},
  year      = {2024},
  pages     = {7036--7050},
  publisher = {Association for Computational Linguistics},
  doi       = {10.18653/v1/2024.naacl-long.389},
  url       = {https://aclanthology.org/2024.naacl-long.389/}
}

@article{cawley2010selection,
 author = {Cawley, Gavin C. and Talbot, Nicola L. C.},
 title = {On Over-fitting in Model Selection and Subsequent Selection Bias in Performance Evaluation},
 journal = {Journal of Machine Learning Research},
 year = {2010}, volume = {11}, number = {70}, pages = {2079--2107},
 url = {https://www.jmlr.org/papers/v11/cawley10a.html}
}

@inproceedings{vanrijn2018importance,
 author = {van Rijn, Jan N. and Hutter, Frank},
 title = {Hyperparameter Importance Across Datasets},
 booktitle = {Proceedings of the 24th ACM SIGKDD International Conference on Knowledge Discovery and Data Mining},
 year = {2018}, pages = {2367--2376}, publisher = {ACM}, doi = {10.1145/3219819.3220058},
 url = {https://arxiv.org/abs/1710.04725}
}

@article{benavoli2017bayesian,
 author = {Benavoli, Alessio and Corani, Giorgio and Dem{\v{s}}ar, Janez and Zaffalon, Marco},
 title = {Time for a Change: a Tutorial for Comparing Multiple Classifiers Through {B}ayesian Analysis},
 journal = {Journal of Machine Learning Research},
 year = {2017}, volume = {18}, number = {77}, pages = {1--36},
 url = {https://jmlr.org/papers/v18/16-305.html}
}

@inproceedings{gulrajani2021domaingeneralization,
 author = {Gulrajani, Ishaan and Lopez-Paz, David},
 title = {In Search of Lost Domain Generalization},
 booktitle = {International Conference on Learning Representations (ICLR)},
 year = {2021},
 url = {https://openreview.net/forum?id=lQdXeXDoWtI}
}

@article{wainer2021nested,
 author = {Wainer, Jacques and Cawley, Gavin},
 title = {Nested Cross-Validation When Selecting Classifiers is Overzealous for Most Practical Applications},
 journal = {Expert Systems with Applications},
 year = {2021}, volume = {182}, pages = {115222},
 doi = {10.1016/j.eswa.2021.115222}
}

@misc{fivetran2026dbt,
 author = {{Fivetran}},
 title = {dbt Packages for {Xero}, {QuickBooks}, {Zuora}, {Stripe}, {Workday}, {Zendesk} and {Jira}},
 year = {2026},
 howpublished = {Apache-2.0 repositories under \url{https://github.com/fivetran}},
 note = {Commits pinned in the task manifest}
}

@article{karl2023multiobjective,
  author  = {Karl, Florian and Pielok, Tobias and Moosbauer, Julia and Pfisterer, Florian and Coors, Stefan and Binder, Martin and Schneider, Lennart and Thomas, Janek and Richter, Jakob and Lang, Michel and Garrido-Merch{\'a}n, Eduardo C. and Branke, Juergen and Bischl, Bernd},
  title   = {Multi-Objective Hyperparameter Optimization in Machine Learning---An Overview},
  journal = {ACM Transactions on Evolutionary Learning and Optimization},
  volume  = {3},
  number  = {4},
  year    = {2023}
}

@article{deb2002nsga2,
  author  = {Deb, Kalyanmoy and Pratap, Amrit and Agarwal, Sameer and Meyarivan, T.},
  title   = {A Fast and Elitist Multiobjective Genetic Algorithm: {NSGA-II}},
  journal = {IEEE Transactions on Evolutionary Computation},
  volume  = {6},
  number  = {2},
  pages   = {182--197},
  year    = {2002}
}

@article{chen2023frugalgpt,
  author  = {Chen, Lingjiao and Zaharia, Matei and Zou, James},
  title   = {{FrugalGPT}: How to Use Large Language Models While Reducing Cost and Improving Performance},
  journal = {Transactions on Machine Learning Research},
  year    = {2024}
}

@inproceedings{ong2024routellm,
  author    = {Ong, Isaac and Almahairi, Amjad and Wu, Vincent and Chiang, Wei-Lin and Wu, Tianhao and Gonzalez, Joseph E. and Kadous, M. Waleed and Stoica, Ion},
  title     = {{RouteLLM}: Learning to Route {LLMs} with Preference Data},
  booktitle = {International Conference on Learning Representations (ICLR)},
  year      = {2025}
}

@inproceedings{yang2023opro,
  author    = {Yang, Chengrun and Wang, Xuezhi and Lu, Yifeng and Liu, Hanxiao and Le, Quoc V. and Zhou, Denny and Chen, Xinyun},
  title     = {Large Language Models as Optimizers},
  booktitle = {International Conference on Learning Representations (ICLR)},
  year      = {2024}
}

@inproceedings{khattab2023dspy,
  author    = {Khattab, Omar and Singhvi, Arnav and Maheshwari, Paridhi and Zhang, Zhiyuan and Santhanam, Keshav and Vardhamanan, Sri and Haq, Saiful and Sharma, Ashutosh and Joshi, Thomas T. and Moazam, Hanna and Miller, Heather and Zaharia, Matei and Potts, Christopher},
  title     = {{DSPy}: Compiling Declarative Language Model Calls into Self-Improving Pipelines},
  booktitle = {International Conference on Learning Representations (ICLR)},
  year      = {2024}
}

@inproceedings{liu2024llambo,
  author    = {Liu, Tennison and Astorga, Nicol{\'a}s and Seedat, Nabeel and van der Schaar, Mihaela},
  title     = {Large Language Models to Enhance {B}ayesian Optimization},
  booktitle = {International Conference on Learning Representations (ICLR)},
  year      = {2024}
}

@inproceedings{zheng2023judging,
  author    = {Zheng, Lianmin and Chiang, Wei-Lin and Sheng, Ying and Zhuang, Siyuan and Wu, Zhanghao and Zhuang, Yonghao and Lin, Zi and Li, Zhuohan and Li, Dacheng and Xing, Eric P. and Zhang, Hao and Gonzalez, Joseph E. and Stoica, Ion},
  title     = {Judging {LLM}-as-a-Judge with {MT-Bench} and {Chatbot Arena}},
  booktitle = {Advances in Neural Information Processing Systems 36, Datasets and Benchmarks Track},
  year      = {2023}
}

@inproceedings{wang2022selfconsistency,
  author    = {Wang, Xuezhi and Wei, Jason and Schuurmans, Dale and Le, Quoc and Chi, Ed and Narang, Sharan and Chowdhery, Aakanksha and Zhou, Denny},
  title     = {Self-Consistency Improves Chain of Thought Reasoning in Language Models},
  booktitle = {International Conference on Learning Representations (ICLR)},
  year      = {2023}
}

@misc{parciak2024schemamatching,
  author        = {Parciak, Marcel and Vandevoort, Brecht and Neven, Frank and Peeters, Liesbet M. and Vansummeren, Stijn},
  title         = {Schema Matching with Large Language Models: an Experimental Study},
  year          = {2024},
  eprint        = {2407.11852},
  archivePrefix = {arXiv}
}

@misc{petelin2025pitfalls,
  author        = {Petelin, Ga{\v{s}}per and Cenikj, Gjorgjina},
  title         = {The Pitfalls of Benchmarking in Algorithm Selection: What We Are Getting Wrong},
  year          = {2025},
  eprint        = {2505.07750},
  archivePrefix = {arXiv}
}

@article{chow1970reject,
  author  = {Chow, C. K.},
  title   = {On Optimum Recognition Error and Reject Tradeoff},
  journal = {IEEE Transactions on Information Theory},
  volume  = {16},
  number  = {1},
  pages   = {41--46},
  year    = {1970}
}

@inproceedings{geifman2017selective,
  author    = {Geifman, Yonatan and El-Yaniv, Ran},
  title     = {Selective Classification for Deep Neural Networks},
  booktitle = {Advances in Neural Information Processing Systems 30},
  year      = {2017}
}

@misc{chen2023usc,
  author        = {Chen, Xinyun and Aksitov, Renat and Alon, Uri and Ren, Jie and Xiao, Kefan and Yin, Pengcheng and Prakash, Sushant and Sutton, Charles and Wang, Xuezhi and Zhou, Denny},
  title         = {Universal Self-Consistency for Large Language Model Generation},
  year          = {2023},
  eprint        = {2311.17311},
  archivePrefix = {arXiv}
}

@inproceedings{jiang2023llmblender,
  author    = {Jiang, Dongfu and Ren, Xiang and Lin, Bill Yuchen},
  title     = {{LLM-Blender}: Ensembling Large Language Models with Pairwise Ranking and Generative Fusion},
  booktitle = {Proceedings of the 61st Annual Meeting of the Association for Computational Linguistics (ACL)},
  year      = {2023}
}

@misc{wang2024moa,
  author        = {Wang, Junlin and Wang, Jue and Athiwaratkun, Ben and Zhang, Ce and Zou, James},
  title         = {Mixture-of-Agents Enhances Large Language Model Capabilities},
  year          = {2024},
  eprint        = {2406.04692},
  archivePrefix = {arXiv}
}

\clearpage
\appendix
\crefalias{section}{appendix}\crefalias{subsection}{appendix}
\section{Settings and search domains}\label{app:parameters}
\Cref{tab:parameter-domains} lists the settings of each stage. The default
appears first in each domain. In planner metadata, \code{d} means descriptions
and \code{s} sample values. In reranker metadata, \code{d}, \code{a} and \code{s}
mean descriptions, aliases and numeric similarity signals. The three score
weights of the reranker are fixed after the search reported in
\Cref{sec:lessons}.
\begingroup\small\setlength{\tabcolsep}{3pt}
\begin{longtable}{@{}L{2.5cm}L{1.1cm}L{3.6cm}L{7.0cm}@{}}
\caption{Parameter definitions, defaults and discrete search domains.}\label{tab:parameter-domains}\\
\toprule Parameter & Default & Search values & What it controls\\\midrule\endfirsthead
\toprule Parameter & Default & Search values & What it controls\\\midrule\endhead
\midrule\multicolumn{4}{@{}l}{\textbf{Table assignment}}\\[2pt]
Assignment calls $L$ & 5 & 5, 1, 3, 7, 9 & Calls that each assign every source table.\\[3pt]
\midrule\multicolumn{4}{@{}l}{\textbf{Planning}}\\[2pt]
Maximum plans & 3 & 3, 1, 2, 5 & Proposals returned for each source column.\\[3pt]
Columns per table & 40 & 40, 20, 80 & Maximum columns shown from each candidate table.\\[3pt]
Planner metadata & ds & ds, none, d, s & Optional descriptions and/or sample values.\\[3pt]
Sample count & 5 & fixed & Maximum values per column when samples are shown.\\[3pt]
\midrule\multicolumn{4}{@{}l}{\textbf{Candidate pool}}\\[2pt]
Pool cap $P$ & 50 & 50, 10, 20, 30 & Maximum entries per matching pass, after shaping.\\[3pt]
Retrieval depth $K$ & 30 & 30, 10, 20 & Hits per enriched name and retriever. Searched with no effect: it only orders the pool (Appendix~\ref{sec:lessons}).\\[3pt]
Cosine cutoff $\tau_c$ & 0.5 & 0.5, 0.6, 0.7 & Minimum dense-search score retained. No effect on the pool.\\[3pt]
BM25 cutoff $\tau_b$ & 1.0 & 1, 2, 3 & Minimum lexical-search score retained. No effect on the pool.\\[3pt]
\midrule\multicolumn{4}{@{}l}{\textbf{Cascade}}\\[2pt]
Threshold $\delta$ & 0.4 & any value (replayed per column) & Search the tables linked by foreign keys at or below this confidence.\\[3pt]
Primary bias $b$ & 0.2 & 0.2, 0.0, 0.1, 0.3 & Add to primary scores for merge ordering only.\\[3pt]
\midrule\multicolumn{4}{@{}l}{\textbf{Reranking}}\\[2pt]
Reranker metadata & das & das, da, ds, d, none & Descriptions (d), aliases (a) and similarity values (s) shown for each candidate.\\[3pt]
Verification pass & on & on, off & A second LLM call that may swap the first answer.\\[3pt]
LLM weight $w$ & 0.6875 & 0.6875, 0.5, 0.8, 0.9, 1.0 & Weight on the LLM score; similarity gets $1-w$.\\[3pt]
Cosine weight $a$ & 0.6 & 0.6, 0.25, 0.5, 0.75, 1.0 & Weight on cosine; word overlap gets $1-a$.\\[3pt]
Similarity scale $\mu$ & 0.5 & 0.5, 0.25, 0.75, 1.0 & Scales similarity within final confidence.\\[3pt]
\bottomrule\end{longtable}

\endgroup

\section{Inputs and instructions of the language models}\label{app:features}
The \emph{schema profile} of a task states whether the task needs
transformations; the numbers of source and target tables and columns; the
number of foreign keys in the target and of target tables with foreign-key
links; the shares of source and target columns with descriptions; the share of
target column names that occur in more than one table; and up to 25 table names
on each side. It includes no label and no result of the task itself.

The \emph{label-free signals} used in \Cref{sec:res-table-step-decider} are 36
numbers computed from the schemas, the pipeline's column embeddings and one
table-assignment call, for example the share of source columns whose most
similar target column lies outside the assigned table. Their definitions are
passed to the decider with the values.

The \emph{label-free examples} used in \Cref{sec:res-table-step-decider} are
built as follows. For each source column, the most similar target column inside
the assigned table and the most similar target column in any table are computed
from the embeddings. The decider receives the share of columns for which the
second lies outside the assigned table, and a seeded sample of eight such
columns: the source column and both target columns, with their descriptions and
cosine similarities.

The instruction of the decider for the table step reads:
\begin{quote}\small
You choose a configuration for an LLM schema-matching pipeline on a new task.
You get the options with a short description of each, tasks already measured
with every option (HR@1 = share of source columns whose top match is right, and
cost per run in USD) with each task's schema profile, the profile of the new
task (no outcomes), and the user's goal. Choose the option that best serves the
goal on the new task. Tasks whose schemas resemble the new one are the best
evidence. Answer with JSON only: \{"option": "<exact option name>", "reason":
"<at most 60 words>"\}
\end{quote}

The instruction of the judge reads:
\begin{quote}\small
You help map the columns of a source database schema to a target schema.

For one source column, a matching pipeline was run in two setups. Each setup
searched a different part of the target schema and returned ranked candidate
target columns, each with a confidence score from the matcher, plus the
matcher's rationale for its top choice.

Why there are several setups: the pipeline can first assign the source table to
one target table and search only there (and in tables linked to it by foreign
keys). This helps when many target tables contain similar-looking columns, but
it cannot find the right column when that column lives in a different target
table than the one assigned. Searching all tables avoids that, but is more
easily misled by look-alike columns in other tables. Confidence scores come
from the same matcher in every setup, but each is relative to the candidates
that setup saw.

Your goal: choose the one target column that this source column's data actually
maps to. You may choose any target column listed by any setup. Reply with JSON
only: \{"choice": "<TABLE.COLUMN exactly as listed>", "reason": "<one or two
sentences>"\}
\end{quote}
The judge's message contains the source column (table, column and
descriptions) and, for each setup in random order, what the setup searched, the
rationale for its top answer, and its five highest-ranked candidates with
confidence, description and data type. The instruction is shown for two setups; it names the number of setups it is given.

\section{Scoring definitions}\label{app:scoring}
A target table is an \emph{accepted table} for a source table when it holds a
correct answer for at least one of the source table's columns. A column whose
correct answer is a transformation has a \emph{complete plan} when at least one
proposal has the correct rule, inputs, parameters and join path. For
\emph{pool recall}, a column counts when a complete correct answer is in the
candidate pool, before the cascade decides whether to search linked tables. For
\emph{routing}, a column counts when its first result is in an accepted table;
a wrong column in the right table is a routing hit and not a correct match. For
\emph{HR@1}, a column counts when its first result is a correct answer. Columns
without a labeled answer are not scored, except for abstention.

The decision metrics are defined for a threshold $\vartheta$. An answered
column with a correct first result is a true positive. An answered column with
a wrong first result, or without any correct answer, is a false positive. An
unanswered column that has a correct answer is a false negative, and an
unanswered column without one is a true negative.

The confidence of a candidate $c$ for source column $q$ is
$\kappa = w\,s_{\rm LLM} + (1-w)\,\mu\,[a\cos(e(q),e(c)) + (1-a)\,\mathrm{Jacc}(q,c)]$,
with the LLM's score $s_{\rm LLM}$, embedding $e$ and token overlap
$\mathrm{Jacc}$; $w=0.6875$, $\mu=0.5$, $a=0.6$.

\section{Decisions per stage and the fixed policies}\label{app:stages}
This appendix gives the setting selected for each stage, the reasons for each
selection, the accuracy--cost curve of each stage, and the fixed policies used
on the ten test tasks.

\Cref{tab:cv} gives the setting that the accuracy-then-cost rule of
\Cref{eq:lexi} selects for each stage on the 35 tasks. Two stages change their
setting at a gain in their stage score, one changes it to save cost at equal
score, and two keep the default.

\begin{table}[htbp]\centering\small
\renewcommand{\arraystretch}{1.25}
\caption{Setting selected for each stage by the accuracy-then-cost rule. The
effect is the change of the stage score on held-out families against the
default, in points; the cost is the change of the stage's own cost against the
default. The stage scores are defined in \Cref{tab:stages}.}
\label{tab:cv}
\begin{tabularx}{\linewidth}{@{}L{2.3cm}YYrr@{}}
\toprule
Stage & Default setting & Selected setting & Effect & Cost\\
\midrule
Table assignment & 5 assignment calls & 1 assignment call & $-0.5$ & $-80\%$\\
Planning & 3 proposals; 40 columns per table; descriptions and samples & 5 proposals; 80 columns per table; no metadata & $+3.4$ & $+20\%$\\
Candidate pool & cap of 50 candidates & unchanged & --- & ---\\
Cascade & search linked tables if confidence $\le 0.40$; bias 0.2 & search linked tables if confidence $\le 0.51$; bias 0.1 & $+3.3$ & $+6\%$\\
Reranking & full metadata; verification call & unchanged & --- & ---\\
\bottomrule
\end{tabularx}
\end{table}

For table assignment, one, three and five calls choose an accepted table in
806, 808 and 809 of 864 assignments (the source tables that have a choice, over
three repeated measurements). The test rates the three equivalent with
probability above 0.99, so the cheapest is selected and four fifths of the
assignment calls are saved.

For planning, showing 80 instead of 40 columns of each table and returning five
proposals raises the share of columns with a complete correct plan by 3.4
points on held-out families, and no family becomes worse. Descriptions and
sample values add cost and little accuracy, so they are omitted, and the
planner tokens rise by about 20\%.

The candidate pool keeps its default, because a cap of 50 gives the highest
pool recall on every task and no smaller cap is within one point on held-out
families. Retrieval depth and the similarity cutoffs do not change the pool at
all.

The cascade is fitted on the 1,602 columns whose table has linked tables. The
tree fitted on these columns splits on the confidence of the first pass in all
25 folds, and features of the schema never replaced it. Searching the linked
tables only when that confidence is at most 0.51 improves routing by 3.3 points
(probability 0.97 of being better than the default). Searching them for every
column gains 3.9 points and is rated equivalent, but needs 39\% more matcher
calls where the selected rule needs 6\%.

Reranking also keeps its default. Several cheaper prompts are within one point
of the full prompt when scored on all tasks, but on held-out families the
cheapest of them loses 1.95 points (probability 0.74 of being worse), so the
full prompt with the verification call is kept.

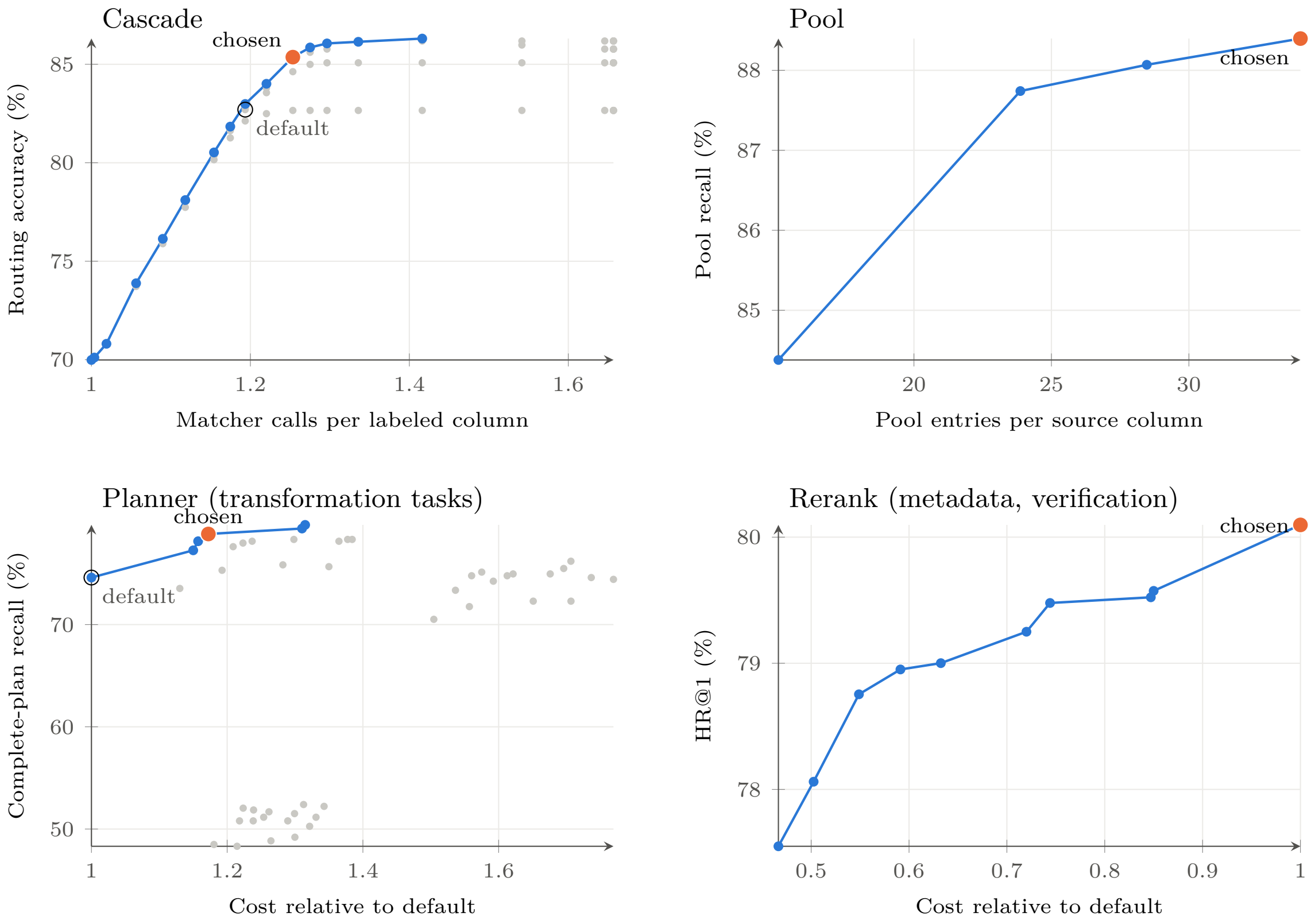
\begin{figure}[htbp]
\centering
\begin{tikzpicture}
\begin{groupplot}[group style={group size=2 by 2, horizontal sep=1.9cm, vertical sep=1.9cm}, width=0.46\textwidth, clip=false, height=0.32\textwidth, tick label style={font=\scriptsize, color=chartmuted}, label style={font=\scriptsize}, title style={font=\small, align=left}, axis lines=left, axis line style={chartmuted}, grid=major, grid style={chartgrid}, every axis title/.style={at={(0,1)}, anchor=south west, font=\small}]
\nextgroupplot[title={Cascade}, xlabel={Matcher calls per labeled column}, ylabel={Routing accuracy (\%)}]
\addplot[only marks, mark=*, mark size=1pt, color=chartother] coordinates {(1.0000,69.988) (1.0037,70.111) (1.0189,70.808) (1.0562,73.883) (1.0898,76.097) (1.1181,78.024) (1.1542,80.361) (1.1747,81.632) (1.1935,82.780) (1.2202,83.805) (1.2534,85.117) (1.2751,85.609) (1.2964,85.773) (1.3358,86.142) (1.4162,86.306) (1.5416,85.978) (1.6458,85.773) (1.6568,85.773) (1.6568,85.773) (1.6568,85.773) (1.6568,85.773) (1.0000,69.988) (1.0037,70.111) (1.0189,70.808) (1.0562,73.883) (1.0898,76.138) (1.1181,78.106) (1.1542,80.525) (1.1747,81.837) (1.1935,82.985) (1.2202,84.010) (1.2534,85.363) (1.2751,85.855) (1.2964,86.060) (1.3358,86.142) (1.4162,86.183) (1.5416,86.183) (1.6458,86.183) (1.6568,86.183) (1.6568,86.183) (1.6568,86.183) (1.6568,86.183) (1.0000,69.988) (1.0037,70.111) (1.0189,70.808) (1.0562,73.883) (1.0898,76.097) (1.1181,78.024) (1.1542,80.402) (1.1747,81.673) (1.1935,82.698) (1.2202,83.559) (1.2534,84.625) (1.2751,84.994) (1.2964,85.076) (1.3358,85.076) (1.4162,85.076) (1.5416,85.076) (1.6458,85.076) (1.6568,85.076) (1.6568,85.076) (1.6568,85.076) (1.6568,85.076) (1.0000,69.988) (1.0037,70.111) (1.0189,70.767) (1.0562,73.719) (1.0898,75.892) (1.1181,77.737) (1.1542,80.156) (1.1747,81.263) (1.1935,82.124) (1.2202,82.493) (1.2534,82.657) (1.2751,82.657) (1.2964,82.657) (1.3358,82.657) (1.4162,82.657) (1.5416,82.657) (1.6458,82.657) (1.6568,82.657) (1.6568,82.657) (1.6568,82.657) (1.6568,82.657)};
\addplot[thick, color=chartfront, mark=*, mark size=1.2pt] coordinates {(1.0000,69.988) (1.0037,70.111) (1.0189,70.808) (1.0562,73.883) (1.0898,76.138) (1.1181,78.106) (1.1542,80.525) (1.1747,81.837) (1.1935,82.985) (1.2202,84.010) (1.2534,85.363) (1.2751,85.855) (1.2964,86.060) (1.3358,86.142) (1.4162,86.306)};
\addplot[only marks, mark=o, mark size=2.4pt, color=black] coordinates {(1.1935,82.698)} node[below right, font=\scriptsize, color=chartmuted] {default};
\addplot[only marks, mark=*, mark size=2.6pt, color=chartchosen, mark options={draw=white}] coordinates {(1.2534,85.363)} node[above left, font=\scriptsize, text=black] {chosen};
\nextgroupplot[title={Pool}, xlabel={Pool entries per source column}, ylabel={Pool recall (\%)}]
\addplot[only marks, mark=*, mark size=1pt, color=chartother] coordinates {(15.0705,84.379) (23.8700,87.741) (28.4682,88.069) (34.0562,88.397)};
\addplot[thick, color=chartfront, mark=*, mark size=1.2pt] coordinates {(15.0705,84.379) (23.8700,87.741) (28.4682,88.069) (34.0562,88.397)};
\addplot[only marks, mark=*, mark size=2.6pt, color=chartchosen, mark options={draw=white}] coordinates {(34.0562,88.397)} node[below left, font=\scriptsize, text=black] {chosen};
\nextgroupplot[title={Planner (transformation tasks)}, xlabel={Cost relative to default}, ylabel={Complete-plan recall (\%)}]
\addplot[only marks, mark=*, mark size=1pt, color=chartother] coordinates {(1.2645,48.845) (1.2998,49.201) (1.1805,48.490) (1.2146,48.313) (1.6510,72.291) (1.7066,72.291) (1.5045,70.515) (1.5569,71.758) (1.2821,75.844) (1.3499,75.666) (1.1302,73.535) (1.1925,75.311) (1.2893,50.799) (1.3215,50.266) (1.2182,50.799) (1.2384,50.799) (1.6760,74.956) (1.7365,74.600) (1.5364,73.357) (1.5922,74.245) (1.2983,78.330) (1.3647,78.153) (1.1500,77.265) (1.2088,77.620) (1.2993,51.510) (1.3309,51.155) (1.2236,52.043) (1.2538,51.155) (1.6959,75.488) (1.0000,74.600) (1.5602,74.778) (1.6126,74.778) (1.3103,79.396) (1.3775,78.330) (1.1572,78.153) (1.2233,77.975) (1.3126,52.398) (1.3428,52.220) (1.2389,51.865) (1.2616,51.687) (1.7065,76.199) (1.7692,74.423) (1.5751,75.133) (1.6213,74.956) (1.3149,79.751) (1.3842,78.330) (1.1723,78.863) (1.2368,78.153)};
\addplot[thick, color=chartfront, mark=*, mark size=1.2pt] coordinates {(1.0000,74.600) (1.1500,77.265) (1.1572,78.153) (1.1723,78.863) (1.3103,79.396) (1.3149,79.751)};
\addplot[only marks, mark=o, mark size=2.4pt, color=black] coordinates {(1.0000,74.600)} node[below right, font=\scriptsize, color=chartmuted] {default};
\addplot[only marks, mark=*, mark size=2.6pt, color=chartchosen, mark options={draw=white}] coordinates {(1.1723,78.863)} node[above, font=\scriptsize, text=black] {chosen};
\nextgroupplot[title={Rerank (metadata, verification)}, xlabel={Cost relative to default}, ylabel={HR@1 (\%)}]
\addplot[only marks, mark=*, mark size=1pt, color=chartother] coordinates {(1.0000,80.097) (0.8471,79.523) (0.7441,79.478) (0.5912,78.951) (0.5488,78.754) (0.8500,79.574) (0.7200,79.250) (0.6325,79.001) (0.5025,78.061) (0.4665,77.549)};
\addplot[thick, color=chartfront, mark=*, mark size=1.2pt] coordinates {(0.4665,77.549) (0.5025,78.061) (0.5488,78.754) (0.5912,78.951) (0.6325,79.001) (0.7200,79.250) (0.7441,79.478) (0.8471,79.523) (0.8500,79.574) (1.0000,80.097)};
\addplot[only marks, mark=*, mark size=2.6pt, color=chartchosen, mark options={draw=white}] coordinates {(1.0000,80.097)} node[left, font=\scriptsize, text=black] {chosen};
\end{groupplot}
\end{tikzpicture}
\caption{Accuracy against cost per stage, replayed from saved runs on all tasks
of the stage. Grey: every setting measured. Blue: settings that no other setting
beats on both axes. Orange: the setting chosen by the accuracy-then-cost rule
under held-out cross-validation. Open circle: the earlier default. For
reranking the chosen point is the default, because the cheaper points lose
accuracy on held-out families.}
\label{fig:pareto}
\end{figure}

\Cref{fig:pareto} shows accuracy against cost for the four stages in which cost
changes. For the cascade and the planner, the accuracy-then-cost rule selects the point
after which the curve flattens; for the pool and the reranking it keeps the
default. Both earlier choices for the cascade are dominated: the default, and
a threshold of 0.8 that a selection without cost chooses. We call
the policy of that selection the \emph{accuracy-only policy}; it searches the
linked tables for nearly every column and keeps the default planner (86.2\%
routing at 1.65 calls per column, against 86.3\% at 1.42 calls for a threshold
of 0.7).

Varying the exchange rate $\lambda$ of \Cref{eq:lambda}, with the settings
fitted on all 28 families, gives six distinct policies (\Cref{tab:policies});
the values $\lambda = 0, 0.5, 1$ and $2$ lead to the same settings and form P1.
These six policies are the ones measured on the ten test tasks. They are not
identical to the policy of \Cref{tab:cv}, which applies the tolerance of one
point of \Cref{eq:lexi} and a column-level tree for the cascade, whereas
\Cref{eq:lambda} trades every point of the stage score against cost and selects
one threshold per policy. The held-out runs of \Cref{tab:heldout} repeat the
construction for every fold, so their settings can differ from fold to fold;
they evaluate the construction for a given $\lambda$ and not the fixed policies
P1 to P6. All policies use one assignment call, plan only on transformation
tasks and search linked tables only for columns that have them.

\begin{table}[htbp]\centering\small
\caption{The policies obtained from the exchange rate $\lambda$. Cascade: the
linked tables are searched when the top confidence is at most $\delta$; $b$ is
the primary bias. Planner: proposals, columns per table, metadata.}
\label{tab:policies}
\begin{tabular}{@{}llllll@{}}
\toprule Policy & $\lambda$ & Cascade $\delta$, $b$ & Planner (transformation tasks) & Pool cap & Reranking prompt\\\midrule
P1 & 0--2 & 0.65, 0.0 & 5, 80, descriptions & 50 & full, verification on\\
P2 & 4 & 0.65, 0.0 & 5, 80, descriptions & 50 & descriptions only, verification on\\
P3 & 8 & 0.60, 0.1 & 5, 80, none & 20 & no metadata, verification on\\
P4 & 16 & 0.55, 0.1 & 5, 80, none & 10 & no metadata, verification off\\
P5 & 32 & 0.50, 0.1 & default (3, 40, descriptions and samples) & 10 & no metadata, verification off\\
P6 & 64 & 0.35, 0.1 & default & 10 & no metadata, verification off\\
\bottomrule
\end{tabular}
\end{table}

\Cref{tab:test10} gives the six policies end to end on the ten test tasks. The
policy P2 reaches 74.4\% at \$1.70, while the accuracy-only policy,
which was selected without cost and searches the linked tables for nearly every column,
reaches 74.8\% at \$2.94. Tuning with cost thus obtains the accuracy of tuning
for accuracy alone, within 0.4 points, at 58\% of its cost. On these ten tasks
the policies are also up to 3.6 points more accurate than the default, but the
tasks belong to the families on which the policies were fitted, and the gain is
not confirmed when groups are held out (\Cref{tab:heldout}).

\begin{table}[htbp]\centering\small
\caption{Fixed policies measured end to end on the ten test tasks: cost of one
pass and mean HR@1 over three seeds, with changes against the default. ``Single
policy'' is the one selected by the rule of \Cref{sec:stage-choice}
(\Cref{tab:cv}); ``accuracy-only'' was selected without cost. The last column
marks the policies that no other row beats on both cost and accuracy.}
\label{tab:test10}
\begin{tabular}{@{}lrrrrl@{}}
\toprule Policy & Cost & Cost change & HR@1 & HR@1 change & Non-dominated\\\midrule
P6 & \$0.62 & $-66\%$ & 64.3 & $-6.5$ & yes\\
P5 & \$0.65 & $-64\%$ & 66.6 & $-4.2$ & yes\\
P4 & \$0.66 & $-64\%$ & 68.1 & $-2.7$ & yes\\
P3 & \$1.10 & $-40\%$ & 72.1 & $+1.3$ & yes\\
P2 & \$1.70 & $-7\%$ & 74.4 & $+3.6$ & yes\\
Default & \$1.82 & --- & 70.8 & --- & no\\
Single policy & \$1.95 & $+7\%$ & 73.5 & $+2.7$ & no\\
P1 & \$2.16 & $+18\%$ & 74.5 & $+3.6$ & yes\\
Accuracy-only policy & \$2.94 & $+62\%$ & 74.8 & $+4.0$ & yes\\
\bottomrule
\end{tabular}
\end{table}

\section{Selection of a policy per task by a language model}\label{app:decider}
\subsection{Method}\label{sec:decider}
No policy in $\Pi$ is the most accurate on every task: on MIMIC-III, P2 loses
7.8 points against the default while P3 keeps the default's accuracy at 35\%
less cost, and on Fivetran Zendesk P2 gains 12.4 points. A selector that
chooses per task can therefore improve on every constant choice. With tens of labeled tasks, a selector fitted on task
features is unreliable; the threshold rule reported below was correct on 30 of
the 36 tasks it was fitted on and no better than a constant on new ones. We therefore do not fit a selector. The selection is made
by a language model, the decider, that is given the measured evidence directly
and asked to choose:
\begin{equation}
 \pi_t \;=\; D\bigl(x_t,\; u,\; \Pi,\; E_{-g(t)}\bigr) \in \Pi,
 \qquad
 E_{-g} = \{(x_{t'}, A(\pi,t'), C(\pi,t')) \in E : g(t') \neq g\}.
 \label{eq:decider}
\end{equation}
Here $E$ is the set of measured results, which holds the context of every
labeled task and the accuracy and cost of every policy on it. The input is a
single structured message with four parts: the policies in
$\Pi$, each with a short description of what it does; for every measured task,
its schema profile and the accuracy and cost of every policy on it; the schema
profile of the new task, without any outcome; and the goal $u$ of the user in
one sentence. The schema profile contains label-free facts about the task: the
numbers of tables and columns on both sides, the foreign keys, the coverage of
descriptions and the names of the tables (\Cref{app:features}). We use three
goals, \emph{accuracy} (pay more when it improves accuracy), \emph{balanced}
(pay extra cost only for a clear gain in accuracy on this kind of task) and
\emph{cost} (accept a small loss of accuracy for a large saving). The output is
the name of one policy and a short justification.

Three properties of this design matter. The decider can only return a member
of $\Pi$, so every decision it can make has been measured end to end, and its
accuracy and cost on the labeled tasks are known in advance. Its instruction
states the task and that measured tasks whose schemas resemble the new one are
the most relevant evidence, and it contains neither a rule nor a threshold, so
that nothing in it is fitted to the labeled tasks. And the context can be
extended with any further label-free evidence without changing the procedure.
We use two extensions below: numeric signals computed
from the schemas and their embeddings, and a small sample of concrete examples
drawn from the new task.

The decider is evaluated with every decision held out, as \Cref{eq:decider}
indicates: for a task of group $g$ the evidence contains only tasks of other
groups. The policy it selects is scored with the measured result of that policy
on the held-out task, so the evaluation needs no executions beyond those in
$E$. We compare the decider with the constant choices, with the virtual best,
and with three fixed rules that receive the same evidence: the policy that is
most accurate on the other tasks, the cheapest policy within $\varepsilon$ of
it, and the cheapest policy. For a choice between two settings we also report
the number of tasks on which the selector chose the better one.

\subsection{Selection among the numeric policies on the ten test tasks}\label{sec:res-decider}
The decider (Claude Opus 5.5) selects one of the policies for each of the ten
test tasks. We evaluate it under two protocols. In the first, each task is held
out in turn, and the decider sees the results of the policies on the other nine
tasks. In the second, which is the protocol of \Cref{eq:decider}, the whole
group of the task is held out: for MIMIC-III and Synthea the decider sees only
accounts receivable and the Fivetran tasks, and for a Fivetran task only
MIMIC-III, Synthea and accounts receivable. Under both protocols the policies
themselves were fitted on all 28 families, which include these ten tasks; what
is held out is the decision. \Cref{tab:decider10} compares the decider with
fixed rules that receive the same evidence.

\begin{table}[htbp]\centering\small
\caption{Selecting one policy per task on the ten test tasks, with each task or
each whole group held out from the evidence. Changes are against the default,
whose cost is \$1.82 per pass at 70.8\% mean HR@1. The cost excludes the
decider's own calls (about \$0.03 per decision).}
\label{tab:decider10}
\begin{tabular}{@{}lrrrr@{}}
\toprule
 & \multicolumn{2}{c}{Task held out} & \multicolumn{2}{c}{Group held out}\\
\cmidrule(lr){2-3}\cmidrule(l){4-5}
Selector & Cost & HR@1 & Cost & HR@1\\\midrule
Decider, goal \emph{balanced} & $-19\%$ & $+4.7$ & $+3\%$ & $-0.9$\\
Decider, goal \emph{accuracy} & $-17\%$ & $+3.4$ & $+3\%$ & $-0.9$\\
Decider, goal \emph{cost} & $-61\%$ & $-2.0$ & $-64\%$ & $-2.7$\\
Rule: cheapest within one point of the most accurate & $-7\%$ & $+3.6$ & $-24\%$ & $+0.4$\\
Rule: most accurate on the other tasks & $+31\%$ & $+1.9$ & $+15\%$ & $+1.7$\\
Rule: cheapest & $-66\%$ & $-6.5$ & $-66\%$ & $-6.5$\\
\midrule
Fixed policy P2 & \multicolumn{2}{c}{$-7\%$ \quad $+3.6$} & \multicolumn{2}{c}{$-7\%$ \quad $+3.6$}\\
Accuracy-only policy & \multicolumn{2}{c}{$+62\%$ \quad $+4.0$} & \multicolumn{2}{c}{$+62\%$ \quad $+4.0$}\\
\bottomrule
\end{tabular}
\end{table}

When only the task is held out, the decider with the \emph{balanced} goal is
4.7 points above the default at 19\% less cost. Its justifications rest on a
task of the same group: for MIMIC-III it selected P3 and pointed to Synthea,
which maps into the same target model, and for each Fivetran task it selected
P2 from the results on the other Fivetran tasks. When the whole group is held
out, this evidence is absent and so is the gain. The decider then selects the
default for every Fivetran task, because the only remaining transformation task
is accounts receivable, on which no policy is clearly better than the default,
and it selects P2 for MIMIC-III and Synthea, which costs MIMIC-III 7.8 points.
The result is 0.9 points below the default, and no selector exceeds the fixed
policy P2 under this protocol. The ten test tasks come from only three groups,
so that each held-out decision is made from one or two dissimilar groups. The
30 decisions cost between \$0.78 and \$1.25 in total.

\subsection{Selection of the table step per task}\label{sec:res-table-step-decider}
For the table step on the 41 tasks, we first fitted a rule that turns the step
off when either of two schema signals exceeds a threshold. The rule was right
on 30 of the 36 tasks on which it was fitted. We wrote the rule down, then
built five new tasks and scored it once: it turned the step off for all five
and was right on three, the same as the constant choice. Leave-one-group-out
selection of a single rule over all 41 tasks reaches 83.8\%, and decision trees
and logistic regression over the signals reach 78--80\%.

The decider chooses between \emph{on} and \emph{off} for each task; for every
group it sees the profiles and the measured HR@1 and cost of both settings for
the tasks of the other 14 groups (\Cref{tab:table-step-decider}). From the
schema profile alone it does not improve on \emph{off}. Adding 36 numeric
signals helps a little. The best of the four versions, which were all compared
on these 41 tasks, replaces statistics by concrete examples: for a sample of
eight source columns of the new task whose most similar target column lies
outside the table that would be assigned, it shows the source column, the most
similar column inside the assigned table and the most similar column anywhere,
with their descriptions. With the examples the decider is right on 33 of 41
tasks and reaches 84.9\%. The examples help because the decider can read what a
statistic cannot express: on MIMIC-III, the most similar column lies outside
the assigned table for 52\% of the source columns, which suggests turning the
step off, but the examples show that these are shared keys such as a person
identifier, and the correct answer is in fact outside the assigned table for
only 9\% of the columns. The 41 decisions of this version cost \$28.2, several
times the cost of the pass that they configure.

\begin{table}[htbp]\centering\small
\caption{Table step on or off, selected per task on the 41 tasks, every group
held out. Right calls count the tasks on which the better setting was selected
(a difference below one point counts as right). Decision cost is the total for
the 41 decisions.}
\label{tab:table-step-decider}
\begin{tabular}{@{}lrrr@{}}
\toprule Selector & HR@1 & Right calls & Decision cost\\\midrule
Always on & 75.7 & 11/41 & ---\\
Always off & 83.1 & 31/41 & ---\\
Decider (Opus 5.5), schema profile & 82.8 & 28/41 & \$3.6\\
Decider (Fable 5.2), schema profile & 83.1 & 29/41 & \$7.6\\
Decider (Fable 5.2), profile and 36 label-free signals & 84.0 & 31/41 & \$15.2\\
Decider (Fable 5.2), profile and label-free examples & 84.9 & 33/41 & \$28.2\\
Oracle per task & 87.8 & 41/41 & ---\\
\bottomrule
\end{tabular}
\end{table}

\section{Abstention}\label{app:abstention}
\subsection{Method}\label{sec:abstention}
The last decision is whether to answer at all, since some items have no
correct answer in the target. The pipeline returns its first answer for item
$i$ when its confidence satisfies $\kappa_i \ge \vartheta$ and otherwise
reports that no match exists. The threshold does not alter the ranking, so it
is selected last, on the confidences produced by the policy already chosen, and
all of its values are evaluated by replay of one stored execution. Let TP be
the items that are answered correctly, FP the items that are answered although
the answer is wrong or no correct answer exists, and FN the items that have a
correct answer and are not answered. The threshold maximizes the F1 score of
this decision, $2\,\mathrm{TP}/(2\,\mathrm{TP}+\mathrm{FP}+\mathrm{FN})$, over a
grid from 0 to 1 in steps of 0.05, with ties resolved toward the lowest
threshold, and it is selected on some tasks and scored on others in the manner
of \Cref{sec:cv}.

\subsection{Results}\label{sec:res-abstention}
Two tasks have labeled source columns without a correct answer: MIMIC-III (153
columns with a match, 112 without) and the internal ERP task (379 with, 698
without). The threshold $\vartheta$ is chosen on one and scored on the other
(\Cref{tab:abstention}).

\begin{table}[htbp]\centering\small
\caption{Abstention threshold $\vartheta$, chosen on one task and scored on the other, with the accuracy-only policy. Mean and standard deviation over three seeds.}
\label{tab:abstention}
\begin{tabular}{@{}lrrrr@{}}
\toprule
 & & \multicolumn{3}{c}{Decision F1}\\
\cmidrule(l){3-5}
Held-out task & HR@1 & Never abstain & $\vartheta$ from the other task & Task's own best $\vartheta$\\\midrule
MIMIC-III & $62.7 \pm 1.7$ & $53.2 \pm 1.1$ & $49.2 \pm 1.3$ ($\vartheta=0.65$) & $67.9 \pm 0.4$ ($\vartheta=0.45$)\\
Internal ERP & $98.9 \pm 0.5$ & $51.7 \pm 0.2$ & $81.0 \pm 0.1$ ($\vartheta=0.45$) & $93.4 \pm 0.4$ ($\vartheta=0.65$)\\
\bottomrule
\end{tabular}
\end{table}

Abstention matters where many columns have no match: on the internal ERP task
it raises F1 from 51.7 to 81.0 with the transferred threshold and to 93.4 with
the task's own. The best threshold differs between the two tasks, and the
threshold of the ERP task is too strict for MIMIC-III. We also selected the
reranking setting by F1 instead of HR@1, with $\vartheta$ chosen jointly; on the held-out task this was not better (in a single run, F1 27.1 against 48.1 on MIMIC-III and 78.7 against 80.9 on the ERP task). The differences in F1 come from
$\vartheta$, not from the reranking setting. The ranking stages are therefore selected on HR@1 and $\vartheta$ on F1, separately for each kind of task.

\section{A third setting of the table step}\label{app:hint}
Besides \emph{on} and \emph{off} we measured a third setting, which retrieves
candidates from all target tables and places the columns of the assigned table
first, so that the assignment acts as a hint. As a fixed setting it reaches
85.6\% mean HR@1 on the 41 tasks at \$7.10 per pass (three seeds), 2.5 points
above \emph{off}. Adding it as a third candidate list for the judge changes the
judge's accuracy by 0.2 points on the same saved candidates (91.6\% against
91.4\%; 95\% interval of the difference $-0.2$ to $0.7$), so the judge of the
main text uses two settings.

\section{Further observations}\label{sec:lessons}
The following observations arose during the study. Several concern the conduct of such a study more than the application.

The most consequential observation concerns the level at which a decision is
made. Decision trees over schema features fitted the training families better
than a constant setting and were worse on held-out families, for every stage,
and for some time this suggested that context carried no usable signal. The
signal appeared once each stage was fitted where it can act and at the level at
which it decides. The cascade, learned from 1,602 columns, splits on the
confidence of the column in every fold, whereas the same procedure applied to
28 family averages found nothing. The table step repeats the pattern one level
higher: the oracle per task reaches 87.8\% on the 41 tasks, and a choice per column exceeds it (\Cref{sec:res-judge}). A selector that finds no context may be looking at the wrong
unit.

Cost plays a related role. When cost is not part of the objective, searching
the linked tables for every column is free, and a selector has no reason to
search only some of them; the accuracy-only selection accordingly chose to
search almost always. With cost in the objective, searching only when the first
pass is uncertain is as accurate for one sixth of the additional calls.
Conversely, cost savings that appear in pooled data need to be confirmed on
held-out data. Scored on all tasks, cheaper reranking prompts appear almost
free; on held-out families the cheapest of them loses about two points, and the
selected option changes from fold to fold.

Two observations concern the evidence given to a selector. A threshold rule
over two schema signals that was correct on 30 of 36 tasks was no better than a
constant on five new tasks; with tens of tasks, a fitted threshold separates
the tasks on which it was fitted. Moreover, label-free signals can measure
something other than what they are meant to measure. Our signals for how often
the answer lies outside the assigned table used the nearest column by embedding
as a substitute for the answer. Where many tables share similar keys, the
nearest column is often in another table while the answer is not: this holds
for 52\% against 9\% of the columns on MIMIC-III, and for 32\% against 0\% on
Synthea. Concrete examples allowed the decider to recognize the difference,
which no aggregate number conveyed.

Some settings cannot change the result, and it is worth establishing this
before spending measurements on them. The three weights of the reranking score
were searched over 100 combinations on 35 tasks. No combination was better than
the default by more than run-to-run noise ($\pm 17$ of 2,439 columns), and
weights selected by leave-one-group-out lost 16 columns on the held-out groups.
The score of the model determines the ranking, and the weights mainly change
how often the cascade searches further; they were fixed and removed from the
search. For the same reason, retrieval depth and similarity cutoffs cannot
change a pool that is defined by the assigned table.

The data mattered more than any setting. The Fivetran source tables declare no
foreign keys, so the cascade initially had nothing to search. Adding the keys
used by the joins of the packages raised mean HR@1 on the seven tasks from
50.9\% to 65.7\% under default settings, several times the gain of any tuned
policy. Run-to-run noise is of the same order as many of the effects under
study: with unchanged settings, three runs of one task scored 70.2\%, 70.2\%
and 85.7\% HR@1 because a single table assignment changed. Repeated runs,
comparisons over groups and a region of practical equivalence are therefore
required before a difference of a few points is interpreted as an effect.
Finally, the abstention threshold did not transfer between the two tasks with
labeled non-matches, and selecting the earlier stages by F1 instead of HR@1 did
not help, which indicates that ranking and the decision to answer are best
selected separately.

\section{Tasks}\label{app:tasks}
\Cref{tab:table-step-tasks} lists the 41 tasks with their groups and the
results of the table-step study per task. The 35 tasks without $^\dagger$
are those of the study of the numeric settings.
\begingroup\small\setlength{\tabcolsep}{3.5pt}
\begin{longtable}{@{}L{4.6cm}L{1.9cm}rrrrrrr@{}}
\caption{The 41 tasks of the table-step study: group, schema sizes (source $\to$ target), scored columns and mean HR@1 over three seeds with the table step on, with it off, for the confidence rule and for the judge. $^\dagger$ added for this study. $^\ast$ used to choose the confidence rule's margin.}\label{tab:table-step-tasks}\\
\toprule Task & Group & Tables & Columns & Scored & On & Off & Rule & Judge\\\midrule\endfirsthead
\toprule Task & Group & Tables & Columns & Scored & On & Off & Rule & Judge\\\midrule\endhead
MIMIC-III to OMOP$^\ast$ & OMOP & 26 $\to$ 38 & 298 $\to$ 427 & 153 & 66.0 & 35.7 & \textbf{67.5} & 67.3\\
Synthea to OMOP & OMOP & 12 $\to$ 38 & 111 $\to$ 427 & 38 & 90.3 & 61.4 & \textbf{96.5} & 95.6\\
AR canonical to operational & AR & 6 $\to$ 4 & 98 $\to$ 136 & 98 & \textbf{91.8} & 75.5 & 89.8 & 90.1\\
Fivetran Jira (dbt)$^\ast$ & Fivetran & 8 $\to$ 18 & 208 $\to$ 133 & 49 & 80.3 & 90.5 & 92.5 & \textbf{98.0}\\
Fivetran QuickBooks (dbt)$^\ast$ & Fivetran & 8 $\to$ 39 & 185 $\to$ 860 & 17 & 31.4 & 82.3 & 76.5 & \textbf{84.3}\\
Fivetran Stripe (dbt)$^\ast$ & Fivetran & 7 $\to$ 27 & 285 $\to$ 666 & 84 & 67.9 & 76.6 & 80.5 & \textbf{83.3}\\
Fivetran Workday (dbt)$^\ast$ & Fivetran & 4 $\to$ 23 & 100 $\to$ 479 & 65 & 55.9 & 81.0 & 90.3 & \textbf{95.9}\\
Fivetran Xero (dbt)$^\ast$ & Fivetran & 5 $\to$ 17 & 127 $\to$ 127 & 67 & 63.7 & 85.1 & 86.1 & \textbf{91.5}\\
Fivetran Zendesk (dbt)$^\ast$ & Fivetran & 6 $\to$ 20 & 272 $\to$ 198 & 97 & 76.0 & \textbf{95.2} & 85.9 & 94.2\\
Fivetran Zuora (dbt)$^\ast$ & Fivetran & 6 $\to$ 19 & 259 $\to$ 834 & 86 & 77.9 & 90.7 & 96.1 & \textbf{97.7}\\
RACT MySQL to Sakila & RACT / Sakila & 8 $\to$ 15 & 59 $\to$ 70 & 39 & 82.0 & 90.6 & 92.3 & \textbf{95.7}\\
RACT MySQL to TPC-H & RACT / Sakila & 8 $\to$ 8 & 59 $\to$ 61 & 29 & 81.6 & 90.8 & 88.5 & \textbf{96.5}\\
RACT Oracle to MySQL & RACT / Sakila & 7 $\to$ 8 & 43 $\to$ 59 & 21 & \textbf{82.5} & \textbf{82.5} & 81.0 & 81.0\\
RACT Oracle to Sakila & RACT / Sakila & 7 $\to$ 15 & 43 $\to$ 70 & 17 & 82.3 & 86.3 & 92.2 & \textbf{94.1}\\
RACT Oracle to TPC-H & RACT / Sakila & 7 $\to$ 8 & 43 $\to$ 61 & 23 & 91.3 & 81.2 & 94.2 & \textbf{95.7}\\
RACT TPC-H to Sakila & RACT / Sakila & 8 $\to$ 15 & 61 $\to$ 70 & 25 & 54.7 & 61.3 & 58.7 & \textbf{66.7}\\
Sakila views$^\dagger$ & RACT / Sakila & 7 $\to$ 16 & 42 $\to$ 89 & 32 & 69.8 & 91.7 & 92.7 & \textbf{95.8}\\
SchemaNet IMDb to Sakila & RACT / Sakila & 7 $\to$ 16 & 39 $\to$ 90 & 16 & 81.2 & 85.4 & \textbf{87.5} & \textbf{87.5}\\
INSPIRE AdV addresses & INSPIRE & 1 $\to$ 72 & 24 $\to$ 1527 & 17 & 13.7 & 15.7 & 25.5 & \textbf{39.2}\\
INSPIRE NL road links & INSPIRE & 1 $\to$ 55 & 46 $\to$ 1406 & 5 & 66.7 & 33.3 & 53.3 & \textbf{73.3}\\
INSPIRE Piemonte auaViews & INSPIRE & 97 $\to$ 14 & 687 $\to$ 335 & 52 & 97.4 & 78.8 & 93.6 & \textbf{98.7}\\
SINT Goby & SINT & 15 $\to$ 2 & 218 $\to$ 31 & 218 & 78.6 & 88.5 & 85.9 & \textbf{91.6}\\
SINT airline data & SINT & 10 $\to$ 3 & 132 $\to$ 28 & 132 & 69.7 & 94.7 & 94.4 & \textbf{96.2}\\
SINT horror-film characters & SINT & 4 $\to$ 3 & 40 $\to$ 17 & 40 & 79.2 & \textbf{100.0} & 84.2 & 99.2\\
SINT monuments & SINT & 8 $\to$ 2 & 74 $\to$ 21 & 74 & 89.6 & 93.2 & 93.2 & \textbf{98.2}\\
SINT paintings collection & SINT & 11 $\to$ 3 & 133 $\to$ 32 & 130 & 85.4 & \textbf{91.5} & 88.0 & 91.3\\
SINT park events & SINT & 6 $\to$ 3 & 56 $\to$ 20 & 56 & 76.8 & 95.2 & 94.6 & \textbf{98.2}\\
SINT research articles & SINT & 13 $\to$ 2 & 167 $\to$ 22 & 167 & 83.0 & \textbf{94.0} & 92.0 & 93.4\\
SINT volcanic eruptions & SINT & 10 $\to$ 2 & 138 $\to$ 33 & 138 & 87.4 & 93.5 & 93.5 & \textbf{94.0}\\
iBench Amalgam a1 to a3 & iBench & 15 $\to$ 3 & 101 $\to$ 20 & 44 & 96.2 & 77.3 & 96.2 & \textbf{99.2}\\
iBench ontology fixture & iBench & 37 $\to$ 75 & 186 $\to$ 281 & 186 & 56.5 & 89.1 & 88.9 & \textbf{92.5}\\
CDISC SDTM to ADaM pilot & CDISC & 22 $\to$ 10 & 313 $\to$ 415 & 66 & 95.0 & 87.9 & 94.4 & \textbf{95.5}\\
Inventory / Northwind & Northwind & 1 $\to$ 2 & 46 $\to$ 51 & 28 & 71.4 & \textbf{100.0} & 98.8 & \textbf{100.0}\\
Northwind views$^\dagger$ & Northwind & 16 $\to$ 13 & 92 $\to$ 89 & 78 & 75.2 & 90.2 & 87.2 & \textbf{95.3}\\
Real estate (REII) & Real estate & 1 $\to$ 2 & 19 $\to$ 36 & 7 & 85.7 & \textbf{100.0} & \textbf{100.0} & \textbf{100.0}\\
SchemaNet Bank1 to Bank2 & SchemaNet Bank & 9 $\to$ 9 & 27 $\to$ 36 & 17 & 88.2 & 90.2 & 92.2 & \textbf{98.0}\\
TPC-DI historical correspondences & TPC-DI & 18 $\to$ 18 & 187 $\to$ 229 & 138 & 95.9 & 97.3 & \textbf{99.0} & \textbf{99.0}\\
AdventureWorks OLTP views$^\dagger$ & AdventureWorks & 20 $\to$ 70 & 258 $\to$ 484 & 161 & 40.6 & 92.1 & 92.3 & \textbf{95.7}\\
AdventureWorks dbt marts$^\dagger$ & AdventureWorks & 8 $\to$ 15 & 82 $\to$ 120 & 66 & 47.5 & 88.9 & 87.4 & \textbf{91.4}\\
jaffle\_shop dbt marts$^\dagger$ & jaffle\_shop & 2 $\to$ 3 & 16 $\to$ 11 & 7 & \textbf{100.0} & 81.0 & \textbf{100.0} & \textbf{100.0}\\
Internal ERP migration$^\dagger$ & Internal ERP & 6 $\to$ 5 & 1171 $\to$ 951 & 379 & 99.0 & 89.9 & 98.2 & \textbf{99.1}\\
\midrule Mean over the 41 tasks & & & & 3{,}162 & 75.7 & 83.1 & 87.4 & \textbf{91.5}\\
\bottomrule\end{longtable}

\endgroup

\end{document}